\documentclass[runningheads]{llncs}
\usepackage[T1]{fontenc}
\usepackage{graphicx}
\usepackage{amsmath,amssymb}
\usepackage{color}
\usepackage{comment}
\usepackage{booktabs}
\usepackage{multirow}
\usepackage{listings}
\usepackage{colortbl, array, xcolor}

\definecolor{codegreen}{rgb}{0,0.6,0}
\definecolor{codegray}{rgb}{0.5,0.5,0.5}
\definecolor{codepurple}{rgb}{0.58,0,0.82}
\definecolor{backcolour}{rgb}{0.95,0.95,0.92}

\lstdefinestyle{mystyle}{
    backgroundcolor=\color{backcolour},   
    commentstyle=\color{codegreen},
    keywordstyle=\color{magenta},
    numberstyle=\tiny\color{codegray},
    stringstyle=\color{codepurple},
    basicstyle=\ttfamily\footnotesize,
    breakatwhitespace=false,         
    breaklines=true,                 
    captionpos=b,                    
    keepspaces=true,                 
    numbers=left,                    
    numbersep=5pt,                  
    showspaces=false,                
    showstringspaces=false,
    showtabs=false,                  
    tabsize=2
}

\newcommand{\LN}[1]{{\texttt{LN}}(#1)}
\newcommand{\GELU}[1]{{\texttt{GELU}}(#1)}
\newcommand{\U}{\mathbf{U}}
\newcommand{\X}{\mathbf{X}}
\newcommand{\Y}{\mathbf{Y}}
\newcommand{\Xin}{\mathbf{X}_{\rm input}}
\newcommand{\Xout}{\mathbf{X}_{\rm output}}

\begin{document}
\title{RaftMLP: How Much Can Be Done Without Attention and with Less Spatial Locality?}
\titlerunning{RaftMLP}
%
\author{Yuki Tatsunami\inst{1,2}\orcidID{0000-0002-7889-8143} \and
Masato Taki\inst{1}\orcidID{0000-0002-5375-7862}}
\authorrunning{Y. Tatsunami and M. Taki}
%
\institute{Rikkyo University, Tokyo, Japan \\ \email{\{y.tatsunami, taki\_m\}@rikkyo.ac.jp} \and AnyTech Co., Ltd., Tokyo, Japan}

\maketitle              
\begin{abstract}
For the past ten years, CNN has reigned supreme in the world of computer vision, but recently, Transformer has been on the rise. However, the quadratic computational cost of self-attention has become a serious problem in practice applications. There has been much research on architectures without CNN and self-attention in this context. In particular, MLP-Mixer is a simple architecture designed using MLPs and hit an accuracy comparable to the Vision Transformer. However, the only inductive bias in this architecture is the embedding of tokens. This leaves open the possibility of incorporating a non-convolutional (or non-local) inductive bias into the architecture, so we used two simple ideas to incorporate inductive bias into the MLP-Mixer while taking advantage of its ability to capture global correlations. A way is to divide the token-mixing block vertically and horizontally. Another way is to make spatial correlations denser among some channels of token-mixing. With this approach, we were able to improve the accuracy of the MLP-Mixer while reducing its parameters and computational complexity. The small model that is RaftMLP-S is comparable to the state-of-the-art global MLP-based model in terms of parameters and efficiency per calculation. Our source code is available at \url{https://github.com/okojoalg/raft-mlp}.

\end{abstract}
\begin{figure*}[!htb]
  \centering
    \includegraphics[width=0.99\linewidth]{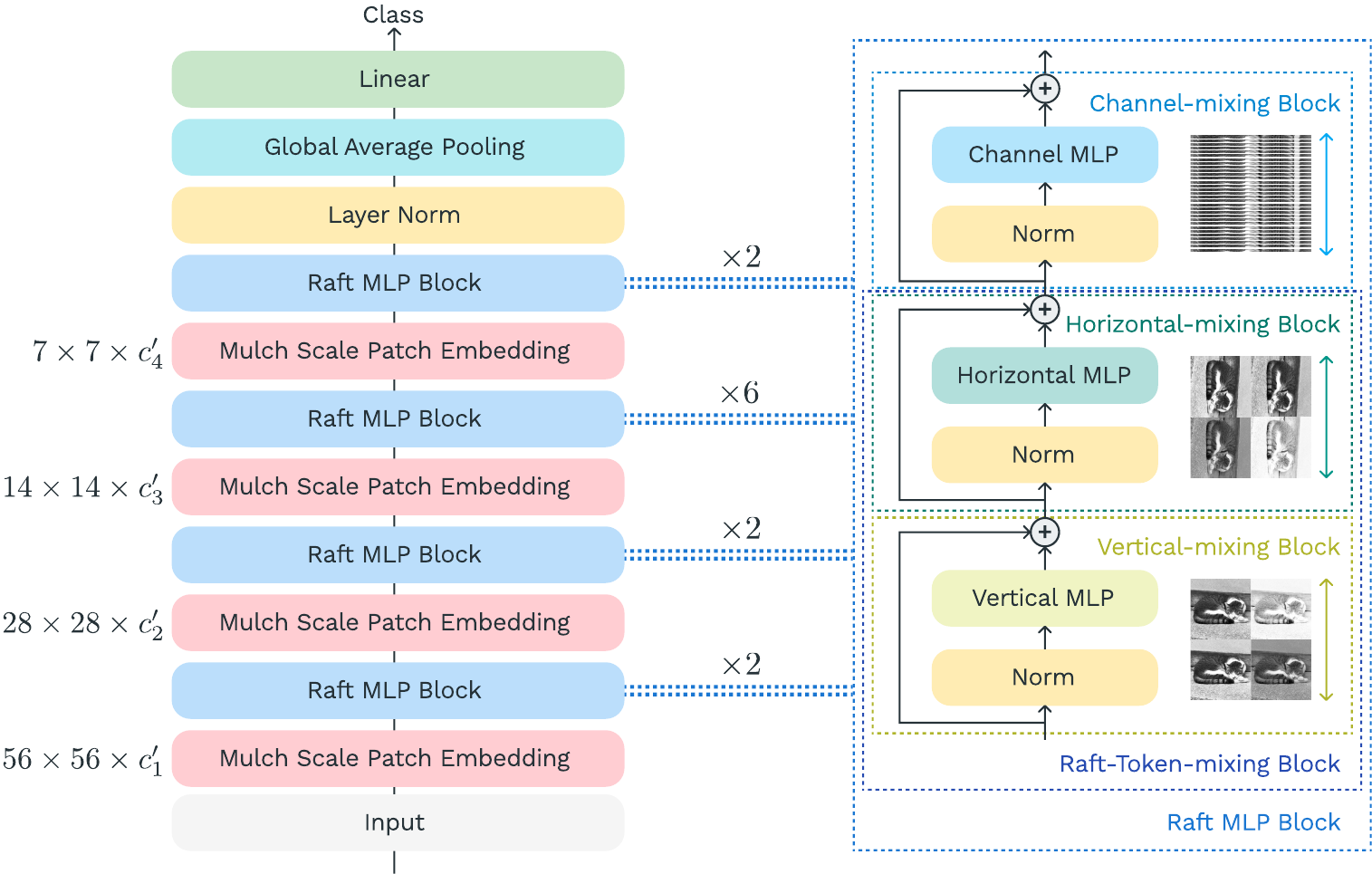}
\caption{The whole architecture of RaftMLP}
\label{fig:raft-mlp}
\end{figure*}

\section{Introduction}
\label{section:introduction}
In the past decade, CNN-based deep architectures have been developed in the computer vision domain. The first of these models was AlexNet~\cite{krizhevsky2012imagenet}, followed by other well-known models such as VGG~\cite{simonyan2014very}, GoogLeNet~\cite{szegedy2015going}, and ResNet~\cite{he2016deep}. These CNN-based models have exhibited high accuracy in various tasks, including image classification, object detection, semantic segmentation, and image generation. Adopting convolution, they employ the inherent inductive bias of images. Meanwhile, Transformer~\cite{vaswani2017attention} has been winning success in recent years in the field of Natural Language Processing (NLP). Inspired by this success, Vision Transformer (ViT)~\cite{dosovitskiy2020image} has been proposed. ViT is a Transformer-based visual model that replaces CNN with the self-attention mechanism. The main idea of ViT is to divide the image into patches based on their spatial locations and apply the Transformer using these patches as tokens. Immediately after the ViT paper appeared, various related works~\cite{arnab2021vivit,chen2021crossvit,ding2021cogview,el2021xcit,han2021transformer,liu2021swin,wang2021pyramid,zhou2021deepvit,yuan2021tokens,zhang2021aggregating} have been done. They have shown that Transformer-based models are competitive with or even exceed CNN-based models in various image recognition and generation tasks. Although Transformer-based models have a reduced inductive bias for images compared to CNN-based models, they compensate for this lack by using a vast array of parameters and computational complexity instead. Moreover, it is successful because it can capture global correlations due to replacing the local receptive fields of convolution with global attention.

More recently, there has been a growing interest in improving the computational complexity of computationally intensive self-attention. Some works~\cite{melas2021you,tolstikhin2021mlp,touvron2021resmlp} claim that Multi Layer Perceptron (MLP) alone is sufficient for image tasks without self-attention. In particular, MLP-Mixer~\cite{tolstikhin2021mlp} has performed a wide variety of MLP-based experiments, and the accuracy of image classification is not better than ViT, but the results are comparable. The MLP-based model, like ViT, first decomposes an image into tokens. A combined operation of MLP, transposition, and activation functions follows the tokenization. The significant point to note is that the transposition operation switches from token-mixing block to channel-mixing block and vice versa. While the channel-mixing block is equivalent to 1x1 convolution in CNN, the token-mixing block is a module that can capture the global correlations between tokens.

The wonderful thing about the MLP-Mixer is that it exhibited the possibility of competing with the existing models with a simple architecture without convolution nor self-attention. In particular, the fact that a simple MLP-based model could compete with current models leads us to think about successors to convolution.
This idea has triggered the interest of many researchers on whether computer vision tasks can outgrow the classical convolution paradigm that has been in the mainstream for ten years. Motivated by the MLP-Mixer, some architectures have been proposed that inject convolutional local structures in pursuit of accuracy. We call the models with such structures local MLP-based models. In contrast, models such as MLP-Mixer, which adopt a design to capture global correlations without local operation, are called global MLP-based models. The global MLP-based model, including MLP-Mixer, has a shortcoming with the models. Unlike convolution, the resolution of the images used for training and inference is fixed, and thwarts the application to downstream tasks such as object detection and semantic segmentation. This paper aims to achieve cost-effectiveness with fewer resources in developing a global MLP-based model. The contributions of this study are as follows.

\paragraph{Spatial structure} As shown in Fig. \ref{fig:raft-mlp}, we propose a module in which the token mixing block is divided into vertical and horizontal mixing blocks in series. In the standard MLP-Mixer, the relevance of patches has no inductive bias in the vertical and horizontal directions in the original two-dimensional image. In our proposed model, we implicitly assume as an inductive bias that patch sequences aligned horizontally have similar correlations with other horizontally aligned patch sequences. The same can be said for vertically aligned patch sequences—additionally, groups of channels are jointed in tensors before inputting into vertical-mixing and horizontal-mixing blocks. Jointed channels are shared with both mixing blocks. Thus, we assume that there are objects and their visual patterns are often distributed linearly over an image and geometrical relation among some channels. 
\paragraph{Multi-scale patch embedding} While ViT and MLP-Mixer patch embedding was a simple method; we added a hierarchical structure. That is multi-scale patch embedding, which embeds information around the patch in the original patch embedding, as shown in Fig. \ref{fig:multi-scale-embedding}. The multi-scale patch embedding method, which also embeds information around the patch in the embedding of the original patch, helped us increase the accuracy at the cost of a small amount of computation and memory consumption.

We will demonstrate that the proposed model with a simple inductive bias without excessive spatial locality as convolution is superior to MLP-Mixer and comparable to global MLP-based models. In addition, we will mention that the proposed method is a model that can achieve accuracy at a reduced cost compared to previous studies. In the appendix, we will study the applicability of the proposed model to downstream tasks such as semantic segmentation, instance segmentation, and object detection. The results will encourage the future possibilities of architectures without self-attention and with less spatial locality.

\section{Related Work}
\label{section:related-work}

\paragraph{Transformer-based models}
Originally proposed for NLP, Transformer~\cite{vaswani2017attention} soon began to be applied to other domains, including visual tasks.
In particular, in image recognition, 
the attention-augmented convolution has been introduced in ~\cite{bello2019attention,hu2018gather,wang2018non}. 
Stand-alone attention for visual task, rather than an augmentation to convolution, is studied in ~\cite{ramachandran2019stand}, where it was shown that fully self-attentional version of ResNet-50 outperforms the original ResNet in ImageNet classification task.

More Transformer-like architectures, process input tokens by self-attention, rather than augmenting CNNs by attention, were studied in ~\cite{cordonnier2019relationship} and ~\cite{dosovitskiy2020image}. In particular, in ~\cite{dosovitskiy2020image}, ViT based on a BERT-type pure Transformer was proposed to deal with high-resolution inputs such as the ImageNet dataset. ViT was pre-trained using a large-scale dataset and transferred to ImageNet, which gave superior results compared to state-of-the-art CNNs.

Inspired by ViT, various transformer-like architectures have been proposed. The most relevant one to our study is CrossFormer~\cite{wang2021crossformer}, which includes a hierarchical structure and Cross-scale Embedding for patch embedding at each level. Cross-scale Embedding effectively injects inductive biases for image domain by using convolution with multiple kernel sizes to perform patch embedding, and it resembles our proposed Multi-scale Patch Embedding in the basic idea. In addition, CrossFormer also proposes a method called Long Short Distance Attention, in which self-attention is divided into two parts, one for long-distance and one for short-distance.

\paragraph{Grobal MLP-based models}
Recently, several alternatives to CNN-based architectures have been proposed that are
simple, yet competitive with CNN despite not using convolution or self-attention ~\cite{tolstikhin2021mlp,melas2021you,touvron2021resmlp}.
MLP-Mixer~\cite{tolstikhin2021mlp} replaces the self-attention layer of ViT with simple cross-tokens MLP.
Despite its simplicity, MLP-Mixer achieves results that are competitive with ViT.
gMLP~\cite{liu2021pay} which consists of an MLP-based module with multiplicative gating is an alternative to MLP-Mixer, achieves higher accuracy than MLP-Mixer with fewer parameters. Vision Permutator~\cite{hou2021vision} focused on mixing in vertical and horizontal directions like our work. Unlike ours, which employs a serialized structure, the Vision Permutator incorporates a parallelized structure, which results in higher accuracy with fewer parameters than the MLP-Mixer.
sMLP~\cite{tang2021sparse} also shares the idea of decomposing token mixing into vertical and horizontal information mixing. These mixings are performed in parallel and the results are added and output from the module.
Another direction of global mixing is CCS-MLP~\cite{yu2021rethinking} as an example. To achieve translation invariance, CCS-MLP introduces circulant token mixing instead of vanilla token mixing MLP.

\paragraph{Local MLP-based models} Moving to a generic inductive bias like Transformer and MLP has attractive possibilities, but its lack of an inductive bias like convolution means that its pre-training requires vast amounts of data compared to CNNs. In order to achieve good performance without large datasets, MLP-based architectures have been proposed as an alternative to MLPs such as S$^2$-MLP~\cite{yu2021s},
S$^2$-MLPv2~\cite{yu2021sv2},
AS-MLP~\cite{lian2021mlp}, CycleMLP~\cite{chen2021cyclemlp}, and ConvMLP~\cite{li2021convmlp}, which incorporate local structures. Although these models have the name of MLP, their essential motivation is the same as CNN in that they use the local structure of the models to extract patterns efficiently. Hence, we call these MLP-based architectures local MLP-based models. In contrast, architectures that mainly utilize MLPs to capture global correlations, such as MLP-Mixer and our study, are called global MLP-based models.
\section{RaftMLP}
\label{section:method}

In this section, we describe MLP-Mixer on which RaftMLP is based and the method adopted for RaftMLP. 

\subsection{Background}

MLP-Mixer~\cite{tolstikhin2021mlp} splits an inputted image into patches of the same size immediately after input and is followed by MLPs that maintain the patch structure. There are two types of MLP: The first one is the token-mixing block, another is the channel-mixing block. We split an image with height $h$ and width $w$ into tokens with height and width $p$. If $h$ and $w$ are divisible by $p$, 
by viewing this image as a collection of these tokens, 
we can regard the image as an data array of height $h'=h/p$ , width $w'=w/p$ and channel $cp^2$ where $c$ denotes channel of the inputted image. The number of a token is then $s=hw/p^2$. The token-mixing block is map $\mathbb{R}^s\to\mathbb{R}^s$ that acts across axes of a token. In contrast, the channel-mixing block is map $\mathbb{R}^c\to\mathbb{R}^c$ that acts across axes of a channel as well where $c$ is the number of channels. Both blocks contain the same modules: Layer Normalization (LN)~\cite{ba2016layer} for each channel, Gaussian Error Linear Units (GELU)~\cite{hendrycks2016gaussian} and MLP. Concretely, the following equation gives the blocks
\begin{equation}
    \Xout = \Xin + W_2\GELU{W_1\LN{\Xin}},
\end{equation}
where $\Xin$ denotes input tensor, $\Xout$ denotes output tensor, $W_1 \in \mathbb{R}^{a\times ae_a}$, $W_2 \in \mathbb{R}^{ae_a\times a}$ denote matrices of MLP layer, and $e_a$ denotes expansion factor. For simplicity, the bias term in MLP was omitted. In token-mixing block, $a=s$ and in channel-mixing block, $a=c$. Moreover, the token-axis and channel-axis are permuted between both mixings. In this way, MLP-Mixer~\cite{tolstikhin2021mlp} is composed of transposition and two types of mixing blocks.

\subsection{Vertical-mixing and Horizontal-mixing Block}
In the previous subsection, we discussed the token-mixing block. The original token-mixing block does not reflect any two-dimensional structure of an input image, such as height or width direction.
In other words, the inductive bias for images is not included in the token-mixing block. MLP-Mixer~\cite{tolstikhin2021mlp} therefore has no inductive bias for images except for how the first patches are made. We decompose this token-mixing block into two blocks that mix vertical and horizontal axes respectively and incorporate inductive bias for image domain. The following describes our method.

The vertical-mixing block is map $\mathbb{R}^{h'}\to\mathbb{R}^{h'}$ that acts across the vertical axis. Precisely, this map captures correlations along the horizontal axis, utilizing the same MLP along the channel and horizontal dimensions. The map also applies layer normalization for each channel, GELU, and the residual connection. The components of this mixing block are the same as the original token-mixing block.

Similarly, the horizontal-mixing block is map $\mathbb{R}^{w'}\to\mathbb{R}^{w'}$, and shuffle the horizontal axis. The structure is dual, only replacing vertical and horizontal axes. We propose replacing token-mixing with a successive application of vertical-mixing and horizontal-mixing, assuming meaningful correlations along vertical and horizontal directions of 2D images. This structure is shown in Fig. \ref{fig:raft-mlp}. The formula is as follows:
\begin{align}
    \U_{*,j,k} =& \X_{*,j,k} + W_{2, \rm ver}\GELU{W_{1, \rm ver}\LN{\X_{*,j,k}}}, \nonumber \\
    &\forall j=1,\ldots,w',\; \forall k=1,\ldots,c, \\
    \Y_{i,*,k} =& \U_{i,*,k} + W_{2, \rm hor}\GELU{W_{1, \rm hor}\LN{\U_{i,*,k}}}, \nonumber \\
    &\forall i=1,\ldots,h',\; \forall k=1,\ldots,c,
\end{align}
where $W_{1, \rm ver} \in \mathbb{R}^{h'\times h'e},\, W_{2, \rm ver} \in \mathbb{R}^{h'e\times h'}, W_{1, \rm hor} \in \mathbb{R}^{w'\times w'e},$ and $W_{2, \rm hor} \in \mathbb{R}^{w'e\times w'}$ denote MLP weight matrices and $\U, \X,$ and $\Y$ denote feature tensors.

\begin{figure*}[htb]
\centering
\includegraphics[width=0.99\linewidth]{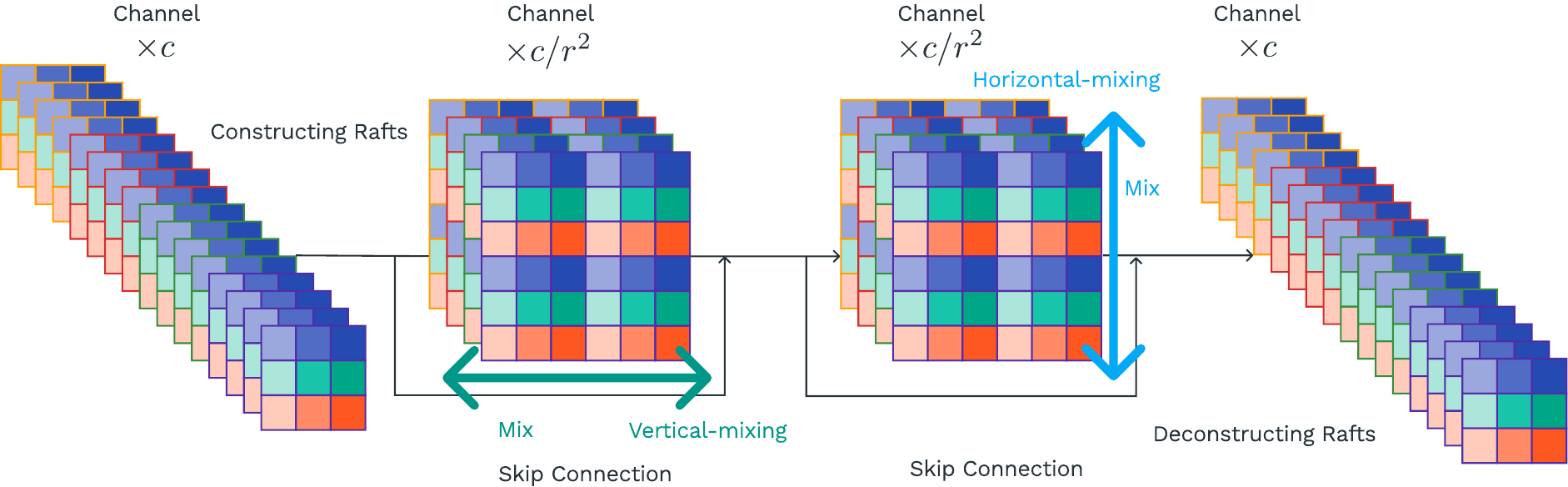}
\caption{The architecture of the raft-token-mixing block. Channels are rearranged with raft-like structure, and then vertical and horizontal mixed.}\label{fig:raft-tmb}
\end{figure*}

\begin{figure}[htb]
\centering
\includegraphics[trim=0 0 0 0,clip, width=0.8\linewidth]{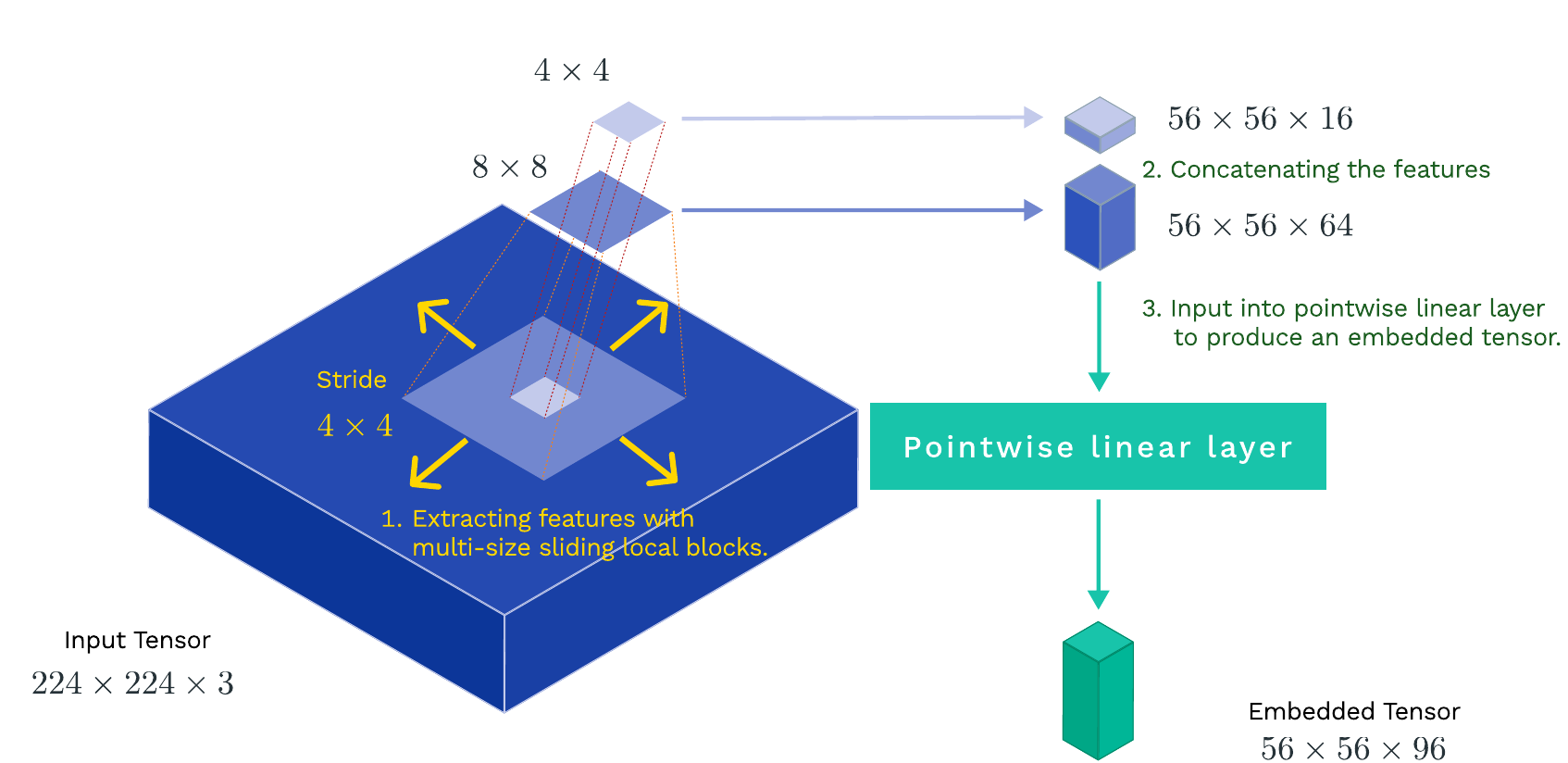}
\caption{A visualization of the concept of multi-scale-embedding.}\label{fig:multi-scale-embedding}
\end{figure}

\subsection{Channel Raft}
Let us assume that several groups of feature map channels have correlations originating from spatial properties. Under this assumption, some feature maps would have some patterns across vertical or horizontal directions. To capture such spatial correlations, we integrate feature maps into the vertical and horizontal shuffle. As shown in Fig. \ref{fig:raft-tmb}, this can be carried out by arranging the feature maps in $h'r \times w'r$, which is reshaping the $h'\times w'\times c$ tensor into a $h'r\times w'r\times c'$ tensor with $c'=c/r^2$ channels. We then perform the vertical-mixing and the horizontal-mixing blocks for this new tensor. In this case, the layer normalization done in each mixing is for the original channel. We refer to this structure as channel raft. The combination of vertical- and horizontal-mixing blocks and the channel raft is called raft-token-mixing block in this paper. The pseudo-code for the raft-token-mixing block is given in Listing \ref{listing:raft-token-mixing}.
The combination of raft-token-mixing block and the channel-mixing block is referred to as RaftMLP block.

\begin{lstlisting}[language=Python, caption=Pseudocode of raft-token-mixing block (Pytorch-like), label=listing:raft-token-mixing]
# b: size of mini -batch, h: height, w: width, 
# c: channel, r: size of raft, o: c//r,
# e: expansion factor,
# x: input tensor of shape (h, w, c)

def __init__(self):
    self.lnv = nn.LayerNorm(c)
    self.lnh = nn.LayerNorm(c)
    self.fnv1 = nn.Linear(r * h, r * h * e)
    self.fnv2 = nn.Linear(r * h * e, r * h)
    self.fnh1 = nn.Linear(r * w, r * w * e)
    self.fnh2 = nn.Linear(r * w * e, r * w)
  
def forward(self, x):
    y = self.lnv(x)
    y = rearrange(y, 'b (h w) (r o) -> b (o w) (r h)')
    y = self.fcv1(y)
    y = F.gelu(y)
    y = self.fcv2(y)
    y = rearrange(y, 'b (o w) (r h) -> b (h w) (r o)')
    y = x + y
    y = self.lnh(y)
    y = rearrange(y, 'b (h w) (r o) -> b (o h) (r w)')
    y = self.fch1(y)
    y = F.gelu(y)
    y = self.fch2(y)
    y = rearrange(y, 'b (o h) (r w) -> b (h w) (r o)')
    return x + y
\end{lstlisting}

\subsection{Multi-scale Patch Embedding}
The majority of both Transformer-based models and MLP-based models are based on patch embedding. We propose an extension of this method named multi-scale patch embedding, which is a patch embedding method that better represents the layered structure of an image. The main idea of the proposed method is twofold. The first is to cut out patches in such a way that the regions overlap. The second is to concatenate the channels of multiple-size patches and then project them by a linear embedding layer.
The outline of the method is shown in Fig. \ref{fig:multi-scale-embedding}, and the details are explained below. First, let $r$ be an arbitrary even number. The method performs zero-padding of $(2^m-1)r/2$ width on the top, bottom, left, and right sides then cut out the patch with $2^mr$ on one side and $r$ stride. In the case of $m=0$, the patch is cut out the same way as in conventional patch embedding. After this patch embedding, the height $h'=h/p$ and width $w'=w/p$ of the tensor is the same, and the output channel is $2^{2m}r^2$. Here, we describe the implementation of multi-scale patch embedding.

Multi-scale patch embedding is a generalization of conventional patch embedding, but it is also slightly different from convolution. However, by injecting a layered structure into the embedding, it can be said to incorporate the inductive bias for images. As the $m$ increases, the computational complexity increases, so we should be careful to decide which $m$ patch cutout to use. Our method is similar to convolutional embedding, but it slightly differs because it uses a linear layer projection after concatenating. See the appendix 
for code details. 

\subsection{Hierarchical Design}
In the proposed method, hierarchical design is introduced. Our architecture used a four-level hierarchical structure with channel raft and multi-scale patch embedding to effectively reduce the number of parameters and improve the accuracy. The hierarchical design is shown in Fig. \ref{fig:raft-mlp}. In this architecture, the number of levels is $L=4$, and at level $l$, after extracting a feature map of $h/2^{l+1}\times w/2^{l+1}\times c_l$ by multi-scale patch embedding, the RaftMLP block is repeated $k_l$ times. The embedding is done using multi-scale patch embedding, but for $l=1,2,3$, the feature maps for $m=0,1$ are concatenated, and for $l=4$, conventional patch embedding is used.
We prepared a hierarchical RaftMLP model with multiple scales. By settling $c'_l$, the number of channels for the level $l$, and  $N_l$, the number of RaftMLP blocks for the level, we developed models for three scales: \textbf{RaftMLP-S}, \textbf{RaftMLP-M}, and \textbf{RaftMLP-L}. The common settings for all three models are vertical dilation expansion factor $e_{\rm ver}=2$, horizontal dilation expansion factor $e_{\rm hor}=2$, channel dilation expansion factor $e_{\rm can}=4$, and channel raft size $r=2$. For patch embedding at each level, multi-scale patch embedding is utilized, but for the $l=1,2,3$ level, patch cutting is performed for $m=0,1$ and then concatenated. For the final level, conventional patch embedding to reduce parameters and computational complexity is utilized. For the output head, a classifier with linear layers and softmax is applied after global average pooling. Refer to
the appendix
for other settings. Our experiments show that the performance of image classification improves as the scale is increased.

\subsection{Impact of Channel Raft on Computational Costs}
We will discuss the computational complexity of channel raft, ignoring normalization and activation functions. Here, let $h'$ denote the height of the patch placement, $w'$ the width of the patch placement, and $e$ the expansion factor.

\paragraph{Number of parameters}
The MLPs parameter for a conventional token-mixing block is 
\begin{equation}
    h'w'(2eh'w'+e+1).
\end{equation}
In contrast, the parameter used for a vertial-mixing block is
\begin{equation}
    h'r(2eh'r+e+1),
\end{equation}
and the parameter used for a holizonal-mixing block is 
\begin{equation}
    w'r(2ew'r+e+1).
\end{equation}
In other words, the total number of parameters required for a raft-token-mixing block is 
\begin{equation}
h'r(2eh'r+e+1)+w'r(2ew'r+e+1).
\end{equation}
This means that if we assume $h'=w'$ and ignore $e+1$, the parameters required for a conventional token-mixing block in the proposed method are $2(r/h')^2$ times for a conventional token-mixing. In short, if we choose $r$ to satisfy $r<h'/\sqrt{2}$, the memory cost can be reduced.

\paragraph{Number of multiply-accumulate}
If we ignore the bias term, the MLPs used for a conventional token-mixing block require $e(h'w')^4$ multiply-accumulates. By contrast, a raft-token-mixing block requires only $er^4(h'^4+w'^4)$. Assuming $h'=w'$, a raft-token-mixing requires only multiply-accumulate of $2r^4/h'^4$ ratio to conventional token-mixing block. To put it plainly, if $r$ is chosen so that $r<h'/2^\frac{1}{4}$, then multiply-accumulation has an advantage over a conventional token-mixing block.

\section{Experimental Evaluation}
\label{section:experimental-evaluation}

\begin{table}[p]
\centering
\caption{Accuracy of the models to be compared with the accuracy of the models derived from the experiments with ImageNet-1k. The throughput measurement infers 16 images per batch using a single V100 GPU. Performance have been not measured for S$^2$-MLP-deep because the code is not publicly available.}
\resizebox{0.65\columnwidth}{!}{
\scalebox{0.9}[0.9]{
\begin{tabular}{@{}clccccr@{}}
\toprule
\multicolumn{1}{c}{\multirow{2}{*}{Backbone}}&\multicolumn{1}{c}{\multirow{2}{*}{Model}}&
\multicolumn{1}{c}{\multirow{1}{*}{\#params}}&
\multicolumn{1}{c}{\multirow{1}{*}{FLOPs}}&
\multicolumn{1}{c}{\multirow{1}{*}{Top-1}}&
\multicolumn{1}{c}{\multirow{1}{*}{Top-5}}&
\multicolumn{1}{c}{\multirow{1}{*}{Throuput}}
\\
&&(M)&(G)&Acc.(\%)&Acc.(\%)&(image/s) \\ \midrule
\multicolumn{6}{l}{\textbf{Low-resource Models}} \\
\multicolumn{6}{l}{(\#params $\times$ FLOPs less than 50P)} \\ \midrule
\multirow{3}{*}{CNN}&ResNet-18~\cite{he2016deep}&11.7&1.8&69.8&89.1&4190\\ 
&MobileNetV3~\cite{Howard2019-ji}&5.4&0.2&75.2&-&1896\\ 
&EfficientNet-B0~\cite{tan2019efficientnet}&5.3&0.4&77.1&-&1275\\ \midrule 
\multirow{2}{*}{Local MLP}&CycleMLP-B1~\cite{chen2021cyclemlp}&15.2&2.1&78.9&-&904\\ 
&ConvMLP-S~\cite{li2021convmlp}&9.0&2.4&76.8&-&1929\\ \midrule 
\multirow{3}{*}{Global MLP}&ResMLP-S12~\cite{touvron2021resmlp}&15.4&3.0&76.6&-&2720\\ 
&gMLP-Ti~\cite{liu2021pay}&6.0&1.4&72.3&-&1194\\ 
\rowcolor[gray]{.9}&RaftMLP-S (\textbf{ours})&9.9&2.1&76.1&93.0&875\\ \midrule 
\multicolumn{6}{l}{\textbf{Middle-Low-resource Models}} \\
\multicolumn{6}{l}{(\#params $\times$ FLOPs more than 50P and less than 150P)} \\ \midrule
\multirow{2}{*}{CNN}&ResNet-50~\cite{he2016deep}&25.6&3.8&76.3&92.2&1652\\ 
&EfficientNet-B4~\cite{tan2019efficientnet}&19.0&4.2&82.6&96.3&465\\ \midrule 
\multirow{5}{*}{Transformer}&DeiT-S~\cite{touvron2020training}&22.1&4.6&81.2&-&1583\\ 
&T2T-ViT$_{t}$-14 ~\cite{yuan2021tokens}&21.5&6.1&81.7&-&849\\ 
&TNT-S~\cite{han2021transformer}&23.8&5.2&81.5&95.7&395\\ 
&CaiT-XS24~\cite{touvron2021going}&26.6&5.4&81.8&-&560\\ 
&Nest-T~\cite{zhang2021aggregating}&17.0&5.8&81.5&-&796\\ \midrule 
\multirow{2}{*}{Local MLP}&AS-MLP-Ti~\cite{lian2021mlp}&28.0&4.4&81.3&-&805\\ 
&ConvMLP-M~\cite{li2021convmlp}&17.4&3.9&79.0&-&1410\\ \midrule 
\multirow{4}{*}{Global MLP}&Mixer-S/16~\cite{tolstikhin2021mlp}&18.5&3.8&73.8&-&2247\\ 
&gMLP-S~\cite{liu2021pay}&19.4&4.5&79.6&-&863\\ 
&ViP-Small/7~\cite{hou2021vision}&25.1&6.9&81.5&-&689\\ 
\rowcolor[gray]{.9}&RaftMLP-M (\textbf{ours})&21.4&4.3&78.8&94.3&758\\ \midrule 
\multicolumn{6}{l}{\textbf{Middle-High-resource Models}} \\
\multicolumn{6}{l}{(\#params $\times$ FLOPs more than 150P and less than 500P)} \\ \midrule
\multirow{3}{*}{CNN}&ResNet-152~\cite{he2016deep}&60.0&11.0&77.8&93.8&548\\ 
&EfficientNet-B5~\cite{tan2019efficientnet}&30.0&9.9&83.7&-&248\\ 
&EfficientNetV2-S~\cite{tan2021efficientnetv2}&22.0&8.8&83.9&-&549\\ \midrule 
\multirow{3}{*}{Transformer}&PVT-M~\cite{wang2021pyramid}&44.2&6.7&81.2&-&742\\ 
&Swin-S~\cite{liu2021swin}&50.0&8.7&83.0&-&559\\ 
&Nest-S~\cite{zhang2021aggregating}&38.0&10.4&83.3&-&521\\ \midrule 
\multirow{4}{*}{Local MLP}&S$^2$-MLP-deep~\cite{yu2021s}&51.0&9.7&80.7&95.4&-\\ 
&CycleMLP-B3~\cite{chen2021cyclemlp}&38.0&6.9&82.4&-&364\\ 
&AS-MLP-S~\cite{lian2021mlp}&50.0&8.5&83.1&-&442\\ 
&ConvMLP-L~\cite{li2021convmlp}&42.7&9.9&80.2&-&928\\ \midrule 
\multirow{3}{*}{Global MLP}&Mixer-B/16~\cite{tolstikhin2021mlp}&59.9&12.6&76.4&-&977\\ 
&ResMLP-S24~\cite{touvron2021resmlp}&30.0&6.0&79.4&-&1415\\ 
\rowcolor[gray]{.9}&RaftMLP-L (\textbf{ours})&36.2&6.5&79.4&94.3&650\\ \midrule 
\multicolumn{6}{l}{\textbf{High-resource Models}} \\
\multicolumn{6}{l}{(Models with \#params $\times$ FLOPs more than 500P)} \\ \midrule
\multirow{4}{*}{Transformer}&ViT-B/16~\cite{dosovitskiy2020image}&86.6&55.5&77.9&-&762\\ 
&DeiT-B~\cite{touvron2020training}&86.6&17.6&81.8&-&789\\ 
&CaiT-S36~\cite{touvron2021going}&68.2&13.9&83.3&-&335\\ 
&Nest-B~\cite{zhang2021aggregating}&68.0&17.9&83.8&-&412\\ \midrule 
\multirow{2}{*}{Global MLP}&gMLP-B~\cite{liu2021pay}&73.1&15.8&81.6&-&498\\ 
&ViP-Medium/7~\cite{hou2021vision}&55.0&16.3&82.7&-&392\\ \bottomrule 
\end{tabular}
}
\label{table:result}
}
\end{table}

In this section, we exhibit experiments for image classification with RaftMLP. In the principal part of this experiment, we utilize the Imagenet-1k dataset~\cite{deng2009imagenet} to train three types of RaftMLP and compare them with MLP-based models and Transformers-based models mainly. We also carry out an ablation study to demonstrate the effectiveness of our proposed method, and as a downstream task, we evaluate transfer learning of RaftMLP for image classification. Besides, We conduct experiments employing RaftMLP as the backbone for object detection and semantic segmentation.

\subsection{ImageNet-1k}
\label{exp-imagenet}
\begin{figure}[!htb]
\centering
\includegraphics[trim=0 0 0 0,clip, width=1.\linewidth]{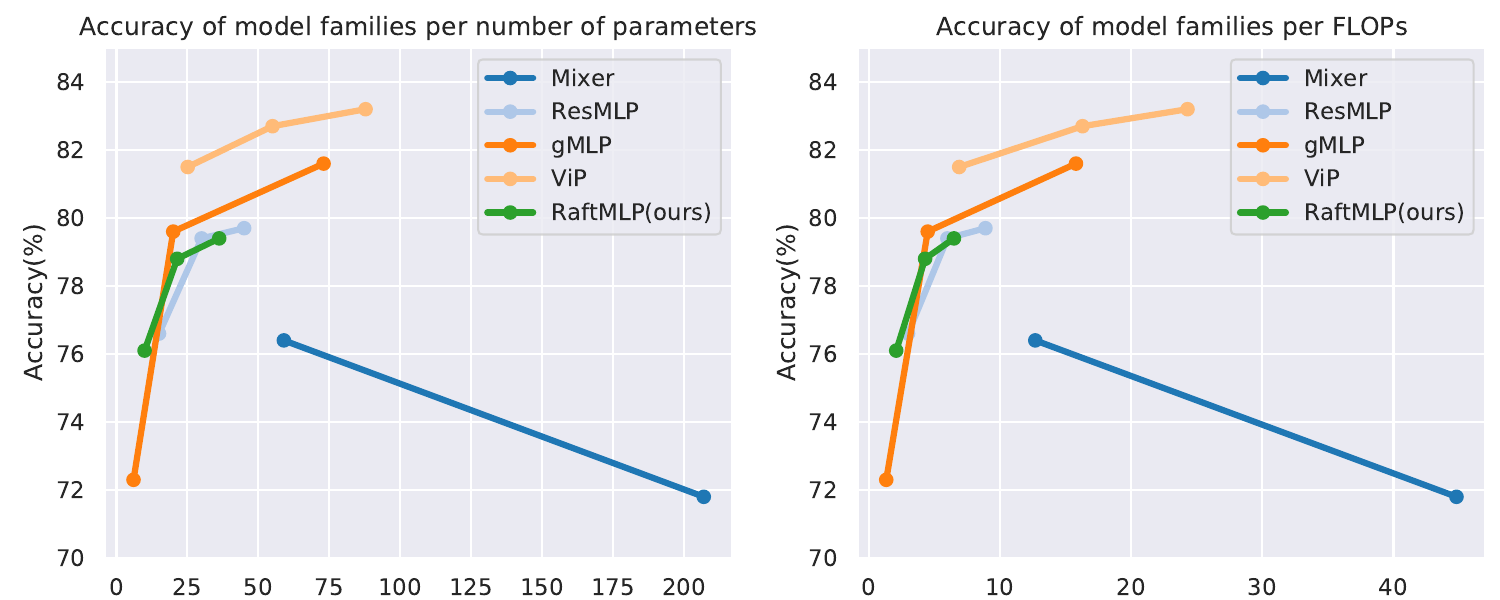}
\caption{Accuracy per parameter and accuracy per FLOPs for the family of global MLP-based models}
\label{fig:acc_p_par_and_flops}
\end{figure}

To evaluate the training results of our proposed classification models, RaftMLP-S, RaftMLP-M and RaftMLP-L, we train them on ImagNet-1k dataset~\cite{deng2009imagenet}. This dataset consists of about 1.2 million training images and about 50,000 validation images assigned 1000 category labels. We also describe how the training is set up below. We employ AdamW~\cite{loshchilov2017decoupled} with weight decay $0.05$ and learning schedule: maximum learning rate $\frac{\rm {batch\, size}}{512} \times 5 \times 10^{-4}$, linear warmup on first $5$ epochs, and after cosine decay to $10^{-5}$ on the following $300$ epochs to train our models. Moreover, we adopt some augmentations and regularizations; random horizontal flip, color jitter, Mixup~\cite{zhang2017mixup} with $\alpha=0.8$, CutMix~\cite{yun2019cutmix} with $\alpha=1.0$, Cutout~\cite{devries2017improved} of rate $0.25$, Rand-Augment~\cite{cubuk2020randaugment}, stochastic depth~\cite{huang2016deep} of rate $0.1$, and label smoothing~\cite{szegedy2016rethinking} $0.1$. These settings refer to the training strategy of DeiT~\cite{touvron2020training}. The other settings are changed for each experiment. Additionally, all training in this experiment is performed on a Linux machine with 8 RTX Quadro 8000 cards. The results of trained models are showed in Table \ref{table:result}. In Fig. \ref{fig:acc_p_par_and_flops}, we compare our method with other global MLP-based models in terms of accuracy against the number of parameters and computational complexity. Fig. \ref{fig:acc_p_par_and_flops} reveals that RaftMLP-S is a cost-effective method.

\subsection{Ablation Study}
\label{exp-ablation}
In order to verify the effectiveness of the two methods we propose, we carry out ablation studies. The setup for these experiments is the same as in Subsection \ref{exp-imagenet}.

\paragraph{Channel Raft (CR)}
We have carried out experiments to verify the effectiveness of channel rafts. Table \ref{table:ablation-channel-raft} compares and verifies MLP-Mixer and MLP-Mixer with the token mixing block replaced by channel rafts. Although we have prepared architectures for $r=1,2,4$ cases, $r=1$ case has no raft structure but is just a conventional token-mixing block vertically and horizontally separated. Table \ref{table:ablation-channel-raft} has shown that channel rafts effectively improve accuracy and costless channel raft structure such as $r=2$ is more efficient for training than increasing $r$.

\begin{table}[htb]
    \centering
    \caption{An ablation experiment of channel raft. Note that Mixer-B/16 is experimented with our implementation}
     \begin{tabular}{crrrr}
        \toprule
        Model & $r$ & \#Mparams & GFLOPs & Top-1 Acc. \\
        \midrule
        Mixer-B/16 & - & 59.9 & 12.6 & 74.3\% \\ \midrule
        \multirow{3}{*}{Mixer-B/16 with CR} & 1 & 58.1 & 11.4 & 77.0\% \\
        & 2 & 58.2 & 11.6 & 78.3\% \\
        & 4 & 58.4 & 12.0 & 78.0\% \\
        \bottomrule
     \end{tabular}
\label{table:ablation-channel-raft}
\end{table}

\paragraph{Multi-scale Patch Embedding (MSPE)}
RaftMLP-M is composed of three multi-scale patch embeddings and a conventional patch embedding. To evaluate the effect of multi-scale patch embedding, we compared RaftMLP-M with the model with multi-scale patch embeddings replaced by conventional patch embeddings in RaftMLP-M. The result is shown on Table \ref{table:ablation-multi-scale}. As a result of comparing the models with and without multi-scale patch embedding, RaftMLP-M with multi-scale patch embedding improves the accuracy by $0.7$\% compared to the model without multi-scale patch embedding.
\begin{table}[htb]
    \centering
    \caption{An ablation experiment of multi-scale patch embedding}
     \begin{tabular}{crrr}
        \toprule
        Model & \#Mparams & GFLOPs & Top-1 Acc. \\
        \midrule
        RaftMLP-M & 21.4 & 4.3 & 78.8\% \\
        RaftMLP-M without MSPE & 20.0 & 3.8 & 78.1\% \\
        \bottomrule
     \end{tabular}
\label{table:ablation-multi-scale}
\end{table}

\subsection{Transfer Learning}
The study of transfer learning is conducted on CIFAR-10/CIFAR-100~\cite{krizhevsky2009learning}, Oxford 102 Flowers~\cite{nilsback2008automated}, Stanford Cars~\cite{krause20133d} and iNaturalist~\cite{van2018inaturalist} to evaluate the transfer capabilities of RaftMLP pre-trained on ImageNet-1k~\cite{deng2009imagenet}. The fine-tuning experiments adopt batch size $256$, weight decay $10^{-4}$ and learning schedule: maximum learning rate $10^{-4}$, linear warmup on first $10$ epochs, and after cosine decay to $10^{-5}$ on the following $40$ epochs. We also do not use stochastic depth~\cite{huang2016deep} and Cutout~\cite{devries2017improved} in this experiment. The rest of the settings are equivalent to Subsection \ref{exp-imagenet}. In our experiments, we also resize all images to the exact resolution $224\times 224$ as ImageNet-1k.
The experiment is shown in Table \ref{table:transfer}. We achieve that RaftMLP-L is more accurate than Mixer-B/16 in all datasets.
\begin{table}[htb]
    \centering
    \caption{The accuracy of transfer learning with each dataset}
     \begin{tabular}{ccccc}
        \toprule
        Dataset & Mixer-B/16& RaftMLP-S & RaftMLP-M & RaftMLP-L \\
        \midrule
        CIFAR-10 & 97.7\% & 97.4\% & 97.7\% & 98.1\% \\
        CIFAR-100 & 85.0\% & 85.1\% & 86.8\% & 86.8\% \\
        Oxford 102 Flowers & 97.8\% & 97.1\% & 97.9\% & 98.4\% \\
        Stanford Cars & 84.3\% & 84.7\% & 87.6\% & 89.0\% \\
        iNaturalist18 & 55.6\% & 56.7\% & 61.7\% & 62.9\% \\
        iNaturalist19 & 64.1\% & 65.4\% & 69.2\% & 70.1\% \\
        \bottomrule
     \end{tabular}
\label{table:transfer}
\end{table}

\section{Discussion}
\label{section:discussion}
The above experimental results show that even an architecture that does not use convolution but has a simple inductive bias for images like vertical and horizontal decomposition can achieve performance competing with Transformers. This is a candidate for minimal inductive biases to improve MLP-based models without convolution. Also, Our method does not require as much computational cost as Transformer. In addition, the computational cost is as expensive as or less than that of CNN. The main reason for the reduced computational cost is that it does not require self-attention. 
The fact that only simple operations such as MLP are needed without self-attention nor convolution means that MLP-based models will be widely used in applied fields since they do not require special software or hardware carefully designed to reduce computational weight.
Furthermore, the raft-token-mixing block has the lead over the token-mixing block of MLP-Mixer in terms of computational complexity when the number of patches is large. As we described in Section \ref{section:method}, substituting the token-mixing block as the raft-token-mixing block reduces parameters from the square of the patches to several times patches. In other words, the more the resolution of images is, the more dramatically parameters are reduced with RaftMLP. The hierarchical design adopted in this paper contributes to the reduction of parameters and computational complexity. Since multi-scale embedding leads to better performance with less cost, our proposal will make it realistic to compose architectures that do not depend on convolution. 
Meanwhile, the experimental results in
the appendix 
suggest that the proposed model 
is not very effective for some downstream tasks. As shown in the appendix, 
the feature map of global MLP-based models differs from the feature map of CNNs in that it is visualized as a different appearance from the input image. Such feature maps are not expected to work entirely in convolution-based architectures such as RetinaNet~\cite{lin2017focal}, Mask R-CNN~\cite{he2017mask}, and Semantic FPN~\cite{kirillov2019panoptic}. Global MLP-based models will require their specialized frameworks for object detection, instance segmentation, and semantic segmentation.

\section{Conclusion}
\label{section:conclusion}
In conclusion, the result has demonstrated that the introduction of the raft-token-mixing block improves accuracy when trained on the ImageNet-1K dataset ~\cite{deng2009imagenet}, as compared to plain MLP-Mixer~\cite{tolstikhin2021mlp}. Although the raft-token-mixing decreases the number of parameters and FLOPs only lightly compared to MLP-Mixer~\cite{tolstikhin2021mlp}, it contributes to the improvement in accuracy in return. We conclude that adding a non-convolutional and non-self-attentional inductive bias to the token-mixing block of MLP-Mixer can improve the accuracy of the model. In addition, due to the introduction of hierarchical structures and multi-scale patch embedding, RaftMLP-S with lower computational complexity and number of parameters have achieved accuracy comparable to the state-of-the-art global MLP-based model with similar computational complexity and number of parameters. We have explicated that it is more cost-effective than the Transformer-based models and well-known CNNs.

However, global MLP-based models have not yet fully explored their potential. Inducing other utilitarian inductive biases, e.g., parallel invariance, may improve the accuracy of global MLP-based models. Further insight into these aspects is left to future work.

\subsubsection{Acknowledgements} We thank the people who support us, belonging to Graduate School of Artificial Intelligence and Science, Rikkyo University.

%
%
%
\bibliographystyle{splncs04}
\bibliography{egbib}

\end{document}


\title{Supplementary Material for "RaftMLP: How Much Can Be Done Without Attention and with Less Spatial Locality?"}
\titlerunning{RaftMLP}
%
\author{Yuki Tatsunami\inst{1,2}\orcidID{0000-0002-7889-8143} \and
Masato Taki\inst{1}\orcidID{0000-0002-5375-7862}}
\authorrunning{Y. Tatsunami and M. Taki}
%
\institute{Rikkyo University, Tokyo, Japan \\ \email{\{y.tatsunami, taki\_m\}@rikkyo.ac.jp} \and AnyTech Co., Ltd., Tokyo, Japan}

\maketitle              
%

%
%
\appendix

\section{More pseudocode}
\label{appendix:pseudocode}
This section describes the pseudo-code for the methods discussed in this paper. The pseudo-code for Multi-scale Patch Embedding is detailed in Listing \ref{listing:multi-scale-patch-embedding}.

\begin{lstlisting}[language=Python, caption=Pseudocode of multi-scale patch embedding (Pytorch-like), label=listing:multi-scale-patch-embedding]
# b: size of mini-batch, h: height, w: width, 
# kernels: list of kernel sizes for unfold.
#          e.g., [4, 8]

def __init__(self, in_channels, out_channels, kernels):
    mlp_in_channels = 0
    for k in kernels:
        mlp_in_channels += k ** 2
    mlp_in_channels *= in_channels
    self.embeddings = nn.ModuleList([
        nn.Sequential(*[nn.Unfold(
                kernel_size=k,
                stride=self.stride,
                padding=(k - self.stride) // 2),
            Rearrange("b c hw -> b hw c")
        ]) for k in kernels
    ])
    self.fc = nn.Linear(
        mlp_in_channels, out_channels
    )

def forward(self, input):
    b, _, h, w = input.shape
    outputs = []
    for emb in self.embeddings:
        output = emb(input)
        outputs.append(output)
    return self.fc(torch.cat(outputs, dim=2))

\end{lstlisting}

\section{Downstream Task Application}
\begin{figure}[!htb]
\centering
\includegraphics[trim=0 0 0 0,clip, width=0.45\linewidth]{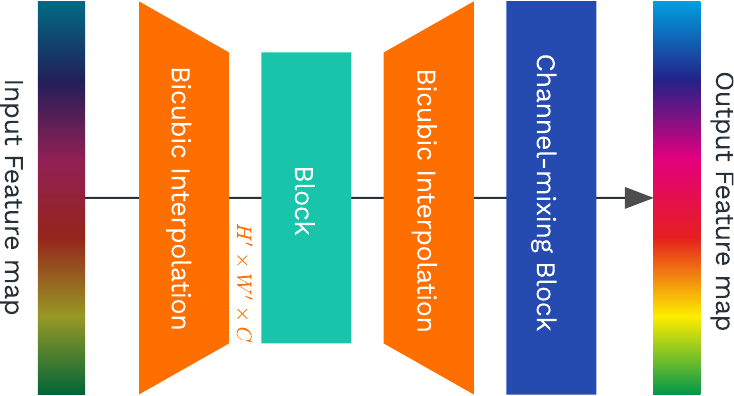}
\caption{Application of RaftMLP block utilizing bicubic interpolation}
\label{fig:bicubic}
\end{figure}

In this section, we discuss the application of our model to fownstream tasks. In order to apply our models to downstream tasks such as semantic segmentation, instance segmentation, and object detection, various resolutions need to be supported. Therefore, we insert bicubic interpolation before and after the raft-token-mixing block, as shown in Fig. \ref{fig:bicubic}. In the bicubic interpolation before the block, we convert the input to the resolution used for pre-training. The bicubic interpolation after the block restores the resolution before the first bicubic interpolation. Moreover, since the resolution of input images is not always divisible by the patch size, we apply bicubic interpolation to obtain the resolution before multi-scale embedding that is a factor of the patch size. This method can be applied to other global MLP-based models such as MLP-Mixer too.

\subsection{Object Detetion}
\label{subsection:detection}
For the evaluation of object detection and instance segmentation, we compose a model in which the backbones of RetinaNet~\cite{lin2017focal} and Mask R-CNN~\cite{he2017mask}, which are both standard implementations on the object detection framework {\tt mmdetection}~\cite{mmdetection}, are replaced by RaftMLP and MLP-Mixer. For the dataset, we used MS COCO~\cite{lin2014microsoft}, which is one of the most popular benchmark datasets for object detection. The training setup is similar to ConvMLP~\cite{li2021convmlp}, with AdamW as the optimizer, learning rate set to $10^{-4}$, weight decay set to $10^{-4}$, and $12$ epochs of training with a batch size of $16$. The results are compared with PureMLP~\cite{li2021convmlp}, ResNet~\cite{he2016deep}, and ConvMLP~\cite{li2021convmlp}, and a summary is provided in Fig. \ref{fig:ap_miou}. See 
Appendix \ref{appendix:quantitive}
for more details.

\subsection{Semantic Segmentation}
\label{subsection:segmentation}
We replace the backbone of Semantic FPN~\cite{kirillov2019panoptic} implemented on {\tt mmsegmention}~\cite{mmseg2020} with RaftMLP and MLP-Mixer and evaluate their performances on the segmentation task. We adopt AdamW as the optimizer with a learning rate of $2.0\times 10^{-4}$ and a weight decay of $10^{-4}$. The learning schedule follows the polynomial decay learning rate policy with a power of 0.9. We use the famous ADE20K dataset~\cite{zhou2017scene} to train the model, with the input image randomly resized and cropped to a resolution of $512\times512$. The model had trained for 40000 iterations. The above settings follow ConvMLP~\cite{li2021convmlp}. A summary of the experimental results is shown in Fig. \ref{fig:ap_miou}. See 
Appendix \ref{appendix:quantitive} for more details.
\begin{figure}[!htb]
\centering
\includegraphics[trim=0 0 0 0,clip, width=1.\linewidth]{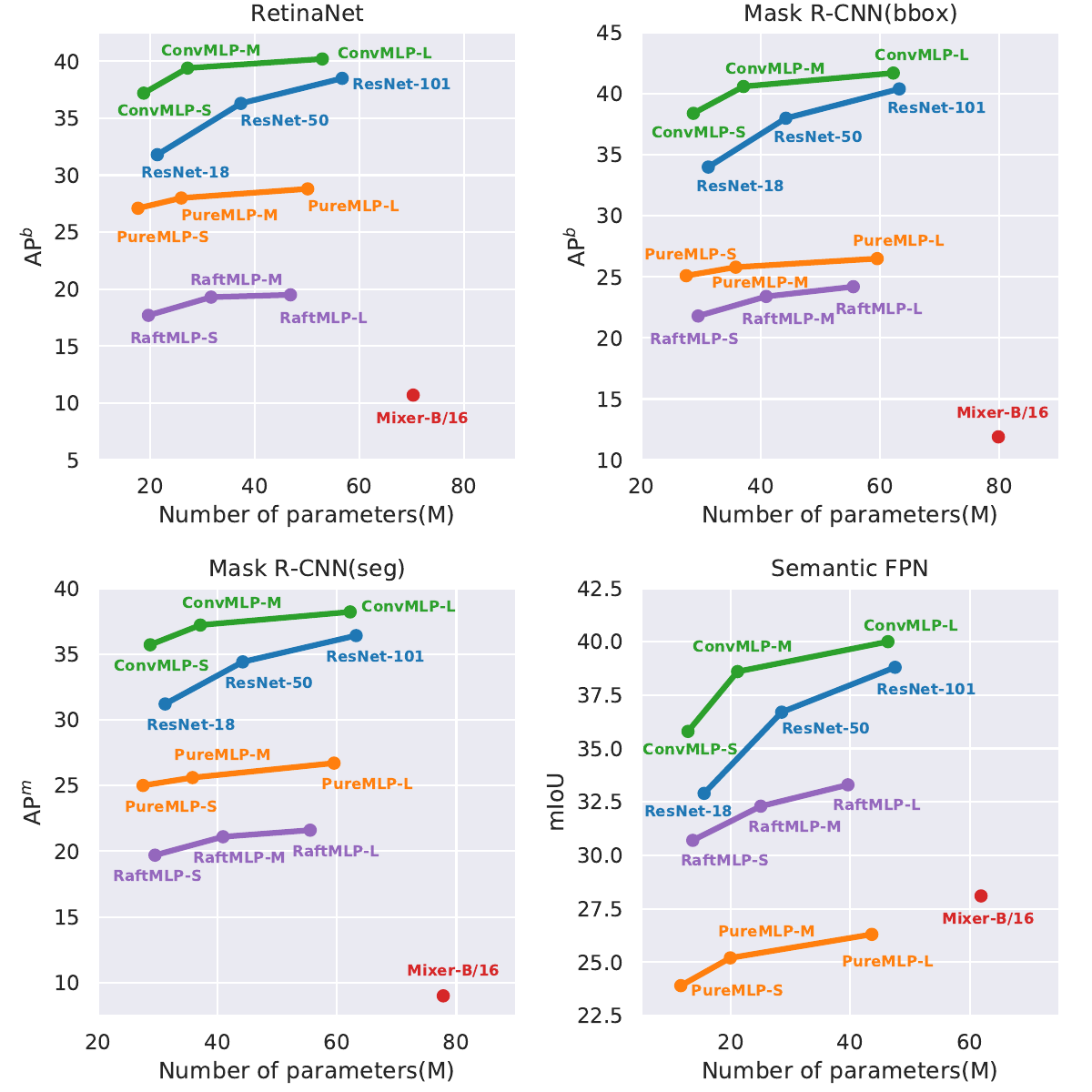}
\caption{The above compares the training results of RetinaNet and Mask R-CNN on MS COCO and Semantic FPN on ADE20K. We compare the results with ResNet, PureMLP, ConvMLP, Mixer, and RaftMLP as backbones in each case. RetinaNet uses AP for bounding boxes, Mask R-CNN AP for bounding boxes and segmentation, and Semantic FPN uses mIoU as their metric.}
\label{fig:ap_miou}
\end{figure}

\subsection{Details of Architectures}
\label{appendix:archtectures}
\paragraph{RaftMLP} The details of the architectures of RaftMLP-S, RaftMLP-M, and RaftMLP-L used in this paper are details in Table \ref{table:models}.
\paragraph{Object Detection, Instance Segmentation and Semantic Segmentation} For the RetinaNet~\cite{lin2017focal} and Mask R-CNN~\cite{he2017mask} and Semantic FPN~\cite{kirillov2019panoptic} we used, we consulted the results of ~\cite{li2021convmlp}, which is the same setup for ResNet, PureMLP, and ConvMLP, which are our comparison. The backbones we have experimented with are RaftMLP and Mixer-B/16. All of the architectures we have arranged use Feature Pyramid Network~\cite{lin2017feature}. Therefore, we must clearly state what feature pyramid was input to these architectures from the backbones RaftMLP and Mixer-B/16. RaftMLP utilizes the output immediately after the first multi-scale patch embedding and the outputs of Level-2 to Level-4 as feature maps to be input to the detector and segmentor. Similarly, Mixer-B/16, along with RaftMLP, uses the output immediately after the patch embedding and the outputs of Block-4, Block-10, and Block-12 as before-mentioned feature maps.

\subsection{Details of Quantitative Results}
\label{appendix:quantitive}
\paragraph{Object Detection and Instance Segmentation} Table \ref{table:retinanet} contains the detailed results of the experiment for RetinaNet performed in Subsection \ref{subsection:detection}, Table \ref{table:maskrcnn} includes the detailed results of the experiment for Mask R-CNN worked in Subsection \ref{subsection:detection}. The results of RetinaNet are not doing as well overall as PureMLP, even with RaftMLP, which guarantees some spatial structure. In particular, it struggles to detect small objects. This result can be seen in \ref{appendix:visualization}, where RaftMLP adds artifacts to the feature map, harming object detection.
\paragraph{Semantic Segmentation} Table \ref{table:semseg} contains the detailed results of the experiment performed in Subsection \ref{subsection:segmentation}.
\subsection{Qualitative Results} 
\paragraph{Object Detection and Instance Segmentation} 
Fig. \ref{fig:detection_gt_image} shows the ground truth for an sample of MS COCO validation dataset. Fig. \ref{fig:retinanet_r50_fpn_1x_coco_image} shows the inference result for RetinaNet with ResNet-50 as the backbone to be installed in {\tt mmdetection}, and Fig. \ref{fig:retinanet_org_mixer_fpn_1x_coco_image}, \ref{fig:retinanet_raftmlp_s_fpn_1x_coco_image}, \ref{fig:retinanet_raftmlp_m_fpn_1x_coco_image}, and \ref{fig:retinanet_raftmlp_l_fpn_1x_coco_image} inference results for the four RetinaNets trained in Subsection \ref{subsection:detection}. Fig. \ref{fig:mask_rcnn_r50_fpn_1x_coco_image} shows the inference result for Mask R-CNN with ResNet-50 as the backbone to be installed in {\tt mmdetection}, and Fig. \ref{fig:maskrcnn_org_mixer_fpn_1x_coco_image}, \ref{fig:maskrcnn_raftmlp_s_fpn_1x_coco_image}, \ref{fig:maskrcnn_raftmlp_m_fpn_1x_coco_image}, and \ref{fig:maskrcnn_raftmlp_l_fpn_1x_coco_image} inference results for the four Mask R-CNNs trained in Subsection \ref{subsection:detection}. Despite the lack of precision, the figures reveal that results of Global MLP-based models for the object detection and instance segmentation tasks are satisfactory.

\begin{figure}[!p]
    \centering
    \begin{tabular}{c}
    \subfloat[Ground Truth]{
        \includegraphics[clip, width=0.4\columnwidth]{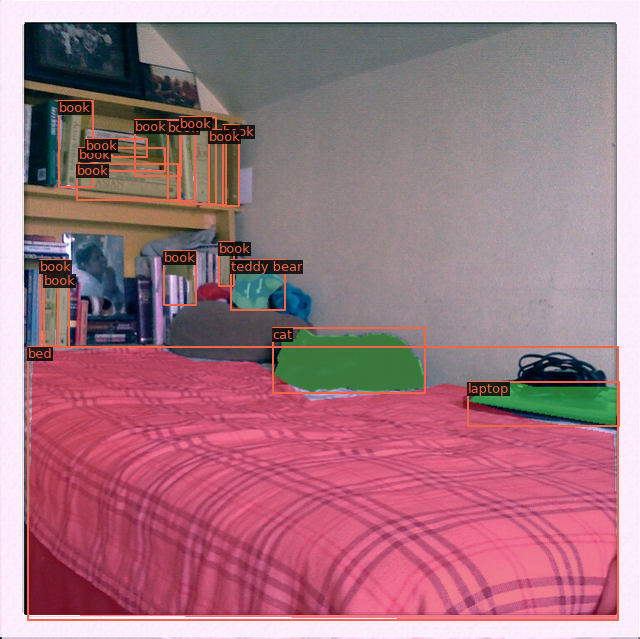}
        \label{fig:detection_gt_image}
    }
    \end{tabular}
    \begin{tabular}{c}
    \subfloat[RetinaNet|ResNet-50]{
        \includegraphics[clip, width=0.4\columnwidth]{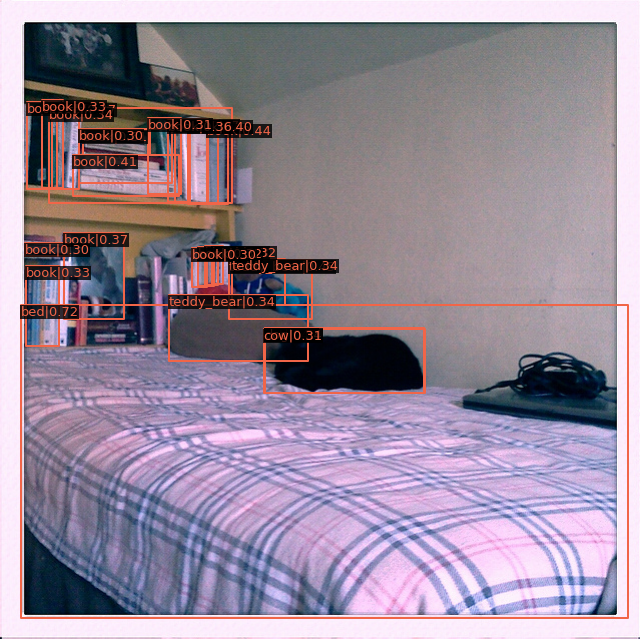}
        \label{fig:retinanet_r50_fpn_1x_coco_image}
    }
    \end{tabular}
    \medskip
    \begin{tabular}{c}
    \subfloat[RetinaNet|Mixer-B/16]{
        \includegraphics[clip, width=0.4\columnwidth]{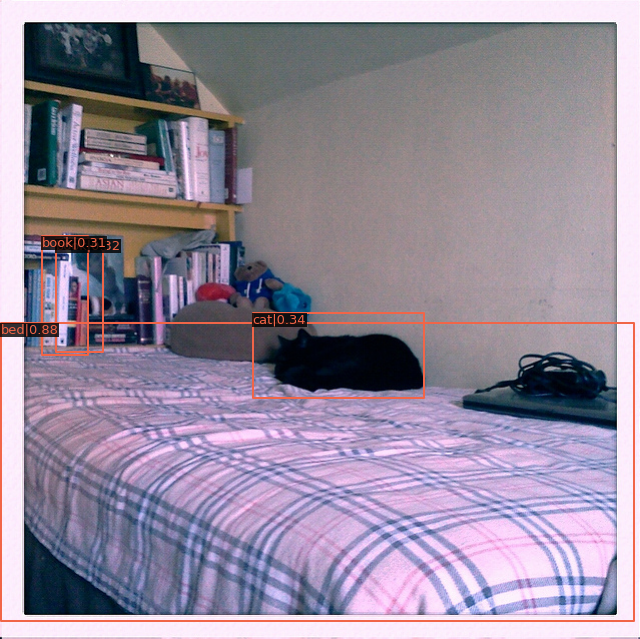}
        \label{fig:retinanet_org_mixer_fpn_1x_coco_image}
    }
    \end{tabular}
    \begin{tabular}{c}
    \subfloat[RetinaNet|RaftMLP-S]{
        \includegraphics[clip, width=0.4\columnwidth]{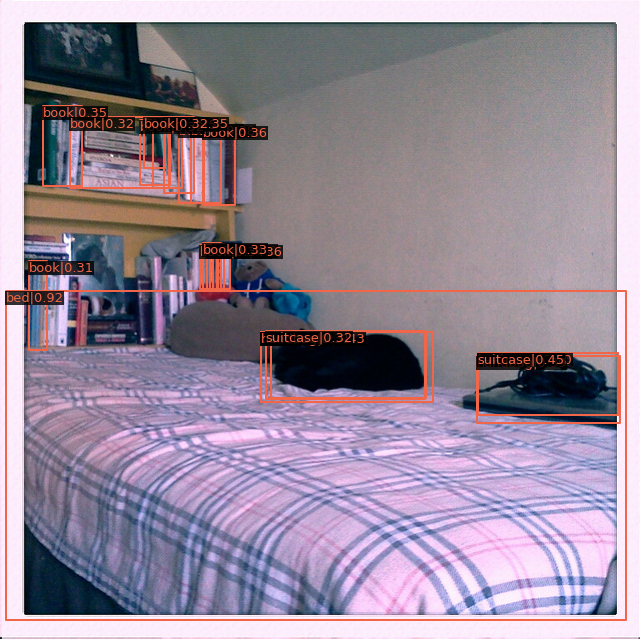}
        \label{fig:retinanet_raftmlp_s_fpn_1x_coco_image}
    }
    \end{tabular}
    \begin{tabular}{c}
    \subfloat[RetinaNet|RaftMLP-M]{
        \includegraphics[clip, width=0.4\columnwidth]{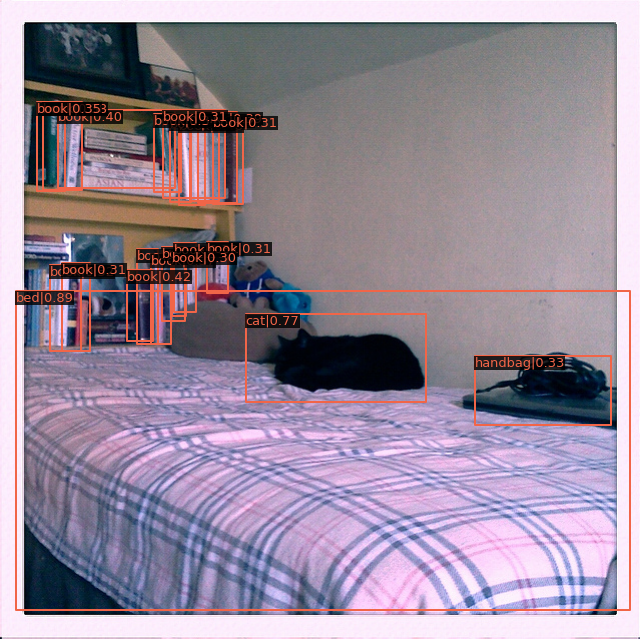}
        \label{fig:retinanet_raftmlp_m_fpn_1x_coco_image}
    }
    \end{tabular}
    \begin{tabular}{c}
    \subfloat[RetinaNet|RaftMLP-L]{
        \includegraphics[clip, width=0.4\columnwidth]{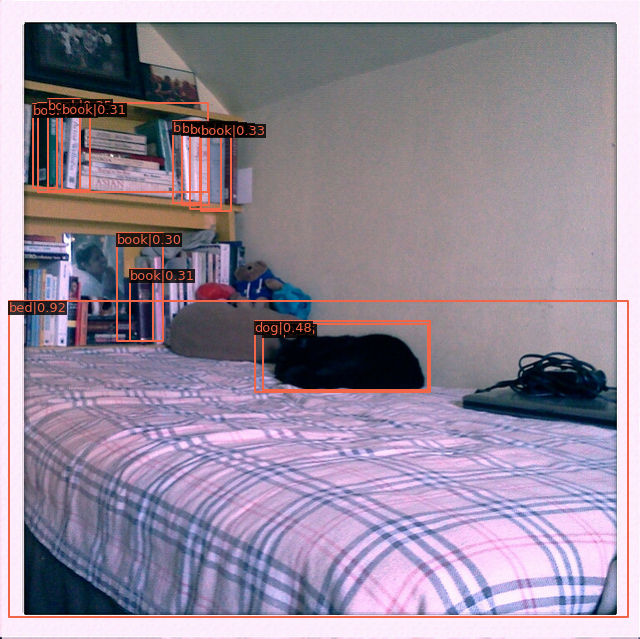}
        \label{fig:retinanet_raftmlp_l_fpn_1x_coco_image}
    }
    \end{tabular}
    \caption{Qualitative results of object detection and instance segmentation}
    \label{fig:detection}
\end{figure}
\begin{figure}[!p]
    \ContinuedFloat
    \centering
    \begin{tabular}{c}
    \subfloat[Mask R-CNN|ResNet-50]{
        \includegraphics[clip, width=0.4\columnwidth]{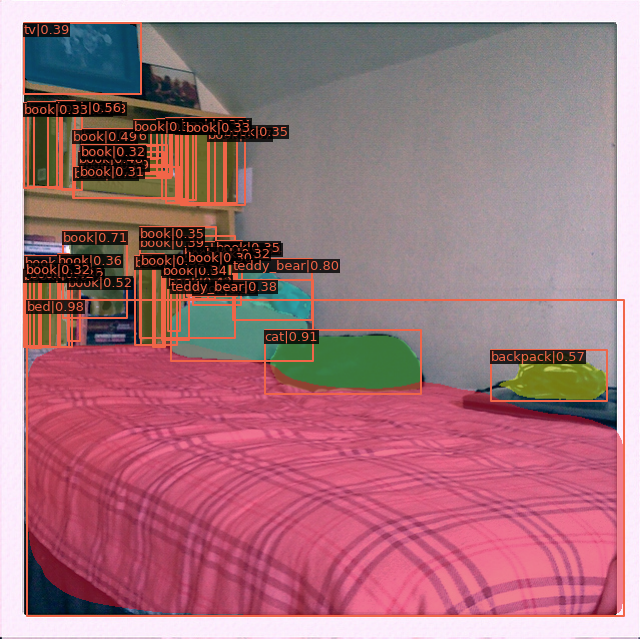}
        \label{fig:mask_rcnn_r50_fpn_1x_coco_image}
    }
    \end{tabular}
    \begin{tabular}{c}
    \subfloat[Mask R-CNN|Mixer-B/16]{
        \includegraphics[clip, width=0.4\columnwidth]{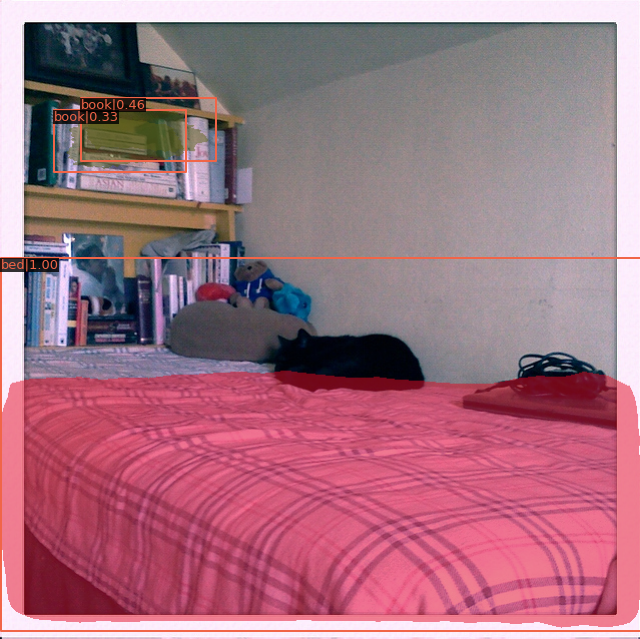}
        \label{fig:maskrcnn_org_mixer_fpn_1x_coco_image}
    }
    \end{tabular}
    \begin{tabular}{c}
    \subfloat[Mask R-CNN|RaftMLP-S]{
        \includegraphics[clip, width=0.4\columnwidth]{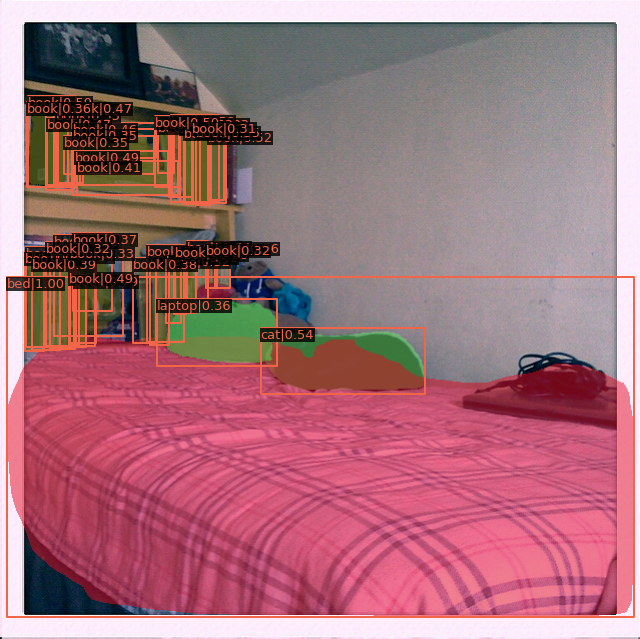}
        \label{fig:maskrcnn_raftmlp_s_fpn_1x_coco_image}
    }
    \end{tabular}
    \begin{tabular}{c}
    \subfloat[Mask R-CNN|RaftMLP-M]{
        \includegraphics[clip, width=0.4\columnwidth]{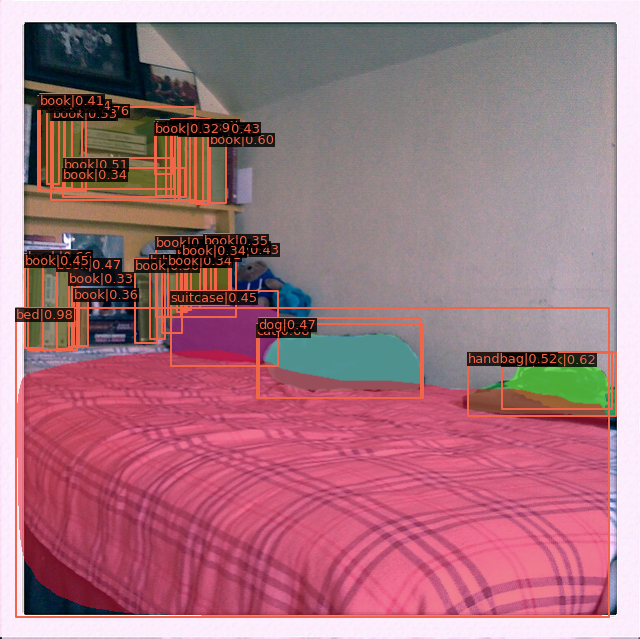}
        \label{fig:maskrcnn_raftmlp_m_fpn_1x_coco_image}
    }
    \end{tabular}
    \begin{tabular}{c}
    \subfloat[Mask R-CNN|RaftMLP-L]{
        \includegraphics[clip, width=0.4\columnwidth]{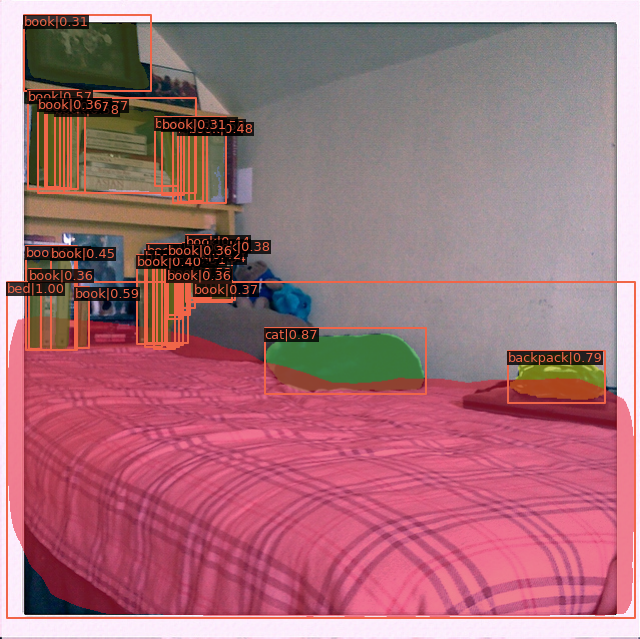}
        \label{fig:maskrcnn_raftmlp_l_fpn_1x_coco_image}
    }
    \end{tabular}
    \caption{Qualitative results of object detection and instance segmentation}
\end{figure}

\begin{figure*}[!ht]
    \centering
    \begin{tabular}{c}
    \subfloat[Ground Truth]{
        \includegraphics[clip, width=0.45\columnwidth]{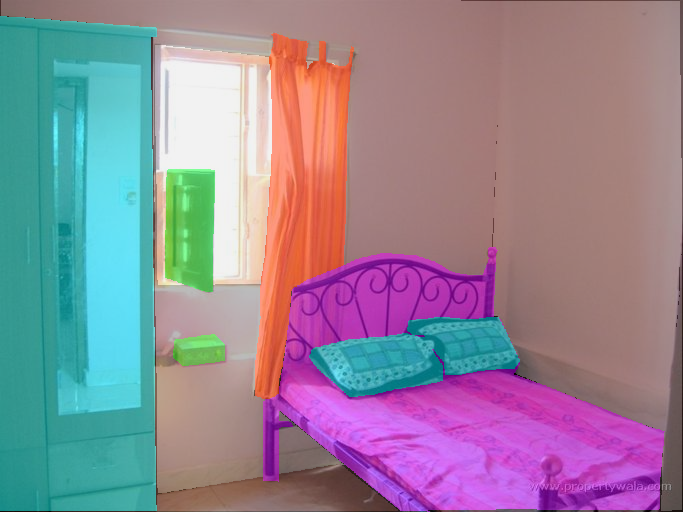}
        \label{fig:segmentation_gt_image}
    }
    \end{tabular}
    \begin{tabular}{c}
    \subfloat[Semantic FPN|ResNet-50]{
        \includegraphics[clip, width=0.45\columnwidth]{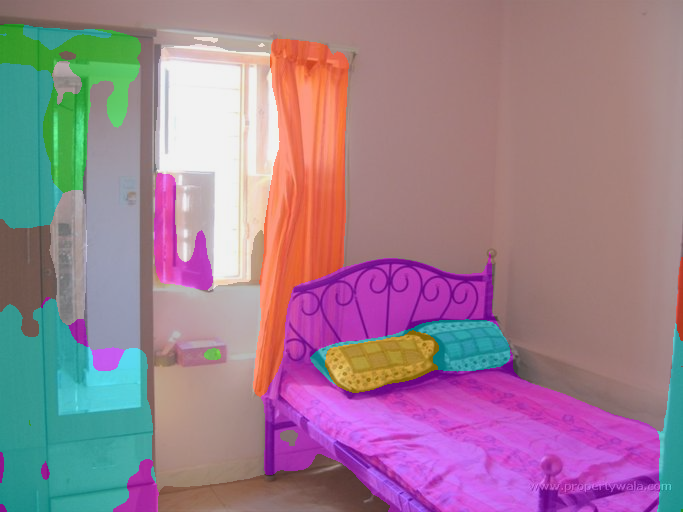}
        \label{fig:fpn_r50_512x512_160k_ade20k_image}
    }
    \end{tabular}
    \begin{tabular}{c}
    \subfloat[Semantic FPN|Mixer-B/16]{
        \includegraphics[clip, width=0.45\columnwidth]{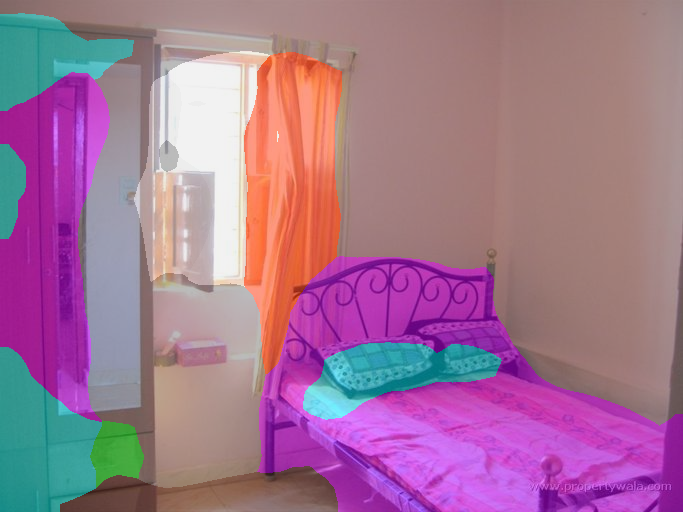}
        \label{fig:fpn_org_mixer_512x512_40k_ade20k_image}
    }
    \end{tabular}
    \begin{tabular}{c}
    \subfloat[Semantic FPN|RaftMLP-S]{
        \includegraphics[clip, width=0.45\columnwidth]{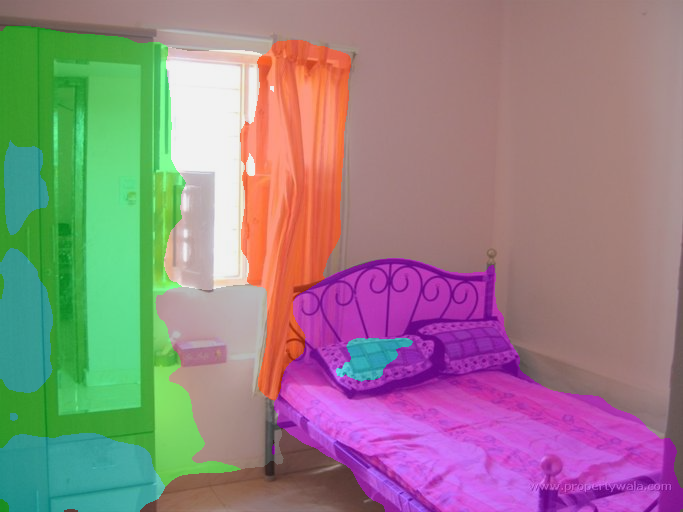}
        \label{fig:fpn_raftmlp_s_512x512_40k_ade20k_image}
    }
    \end{tabular}
    \begin{tabular}{c}
    \subfloat[Semantic FPN|RaftMLP-M]{
        \includegraphics[clip, width=0.45\columnwidth]{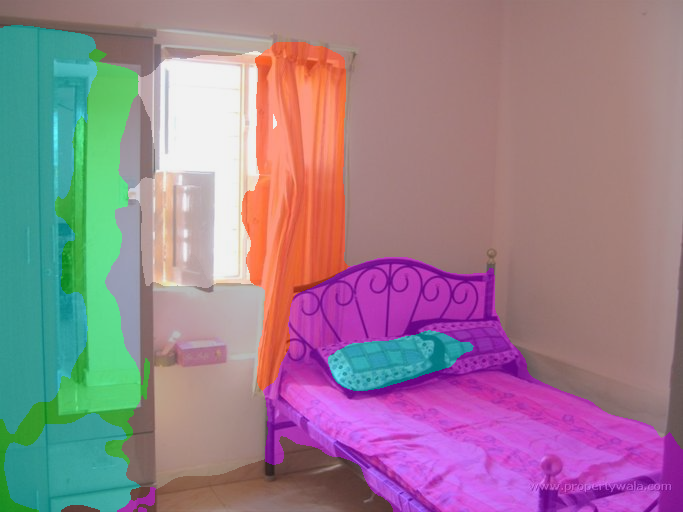}
        \label{fig:fpn_raftmlp_m_512x512_40k_ade20k_image}
    }
    \end{tabular}
    \begin{tabular}{c}
    \subfloat[Semantic FPN|RaftMLP-L]{
        \includegraphics[clip, width=0.45\columnwidth]{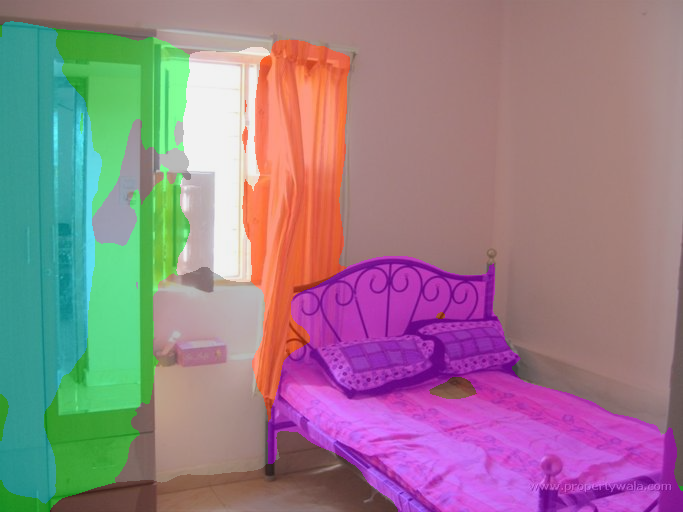}
        \label{fig:fpn_raftmlp_l_512x512_40k_ade20k_image}
    }
    \end{tabular}
    \caption{Qualitative results of semantic segmentation}
    \label{fig:segmentation}
\end{figure*}

\paragraph{Semantic Segmentation} 
Fig. \ref{fig:segmentation_gt_image} presents an image with ADE20k validation dataset overlaid with its ground truth. Fig. \ref{fig:fpn_r50_512x512_160k_ade20k_image} shows the inference results of the model applying ResNet-50, which {\tt mmsegmentation} provides, as the backbone of the Semantic FPN. Fig. \ref{fig:fpn_org_mixer_512x512_40k_ade20k_image}, \ref{fig:fpn_raftmlp_s_512x512_40k_ade20k_image}, \ref{fig:fpn_raftmlp_m_512x512_40k_ade20k_image}, and \ref{fig:fpn_raftmlp_l_512x512_40k_ade20k_image} show the inference results for the four models we trained in Subsection \ref{subsection:segmentation} experiment.

\subsection{Visualization}
\label{appendix:visualization}
\begin{figure*}[!hbt]
\centering
\includegraphics[trim=0 0 0 0,clip, width=0.69\linewidth]{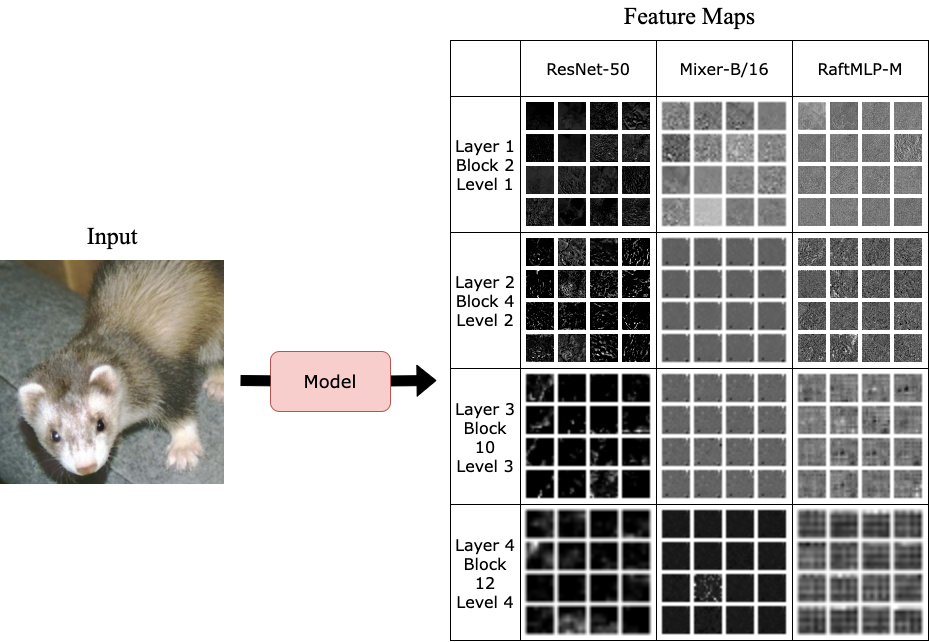}
\caption{Summary of the comparison of ResNet-50, Mixer-B/16, and RaftMLP-M intermediate layer feature maps}
\label{fig:featuremaps}
\end{figure*}

\begin{figure}[htb]
\centering
\includegraphics[trim=0 1045 0 118,clip, width=0.99\linewidth]{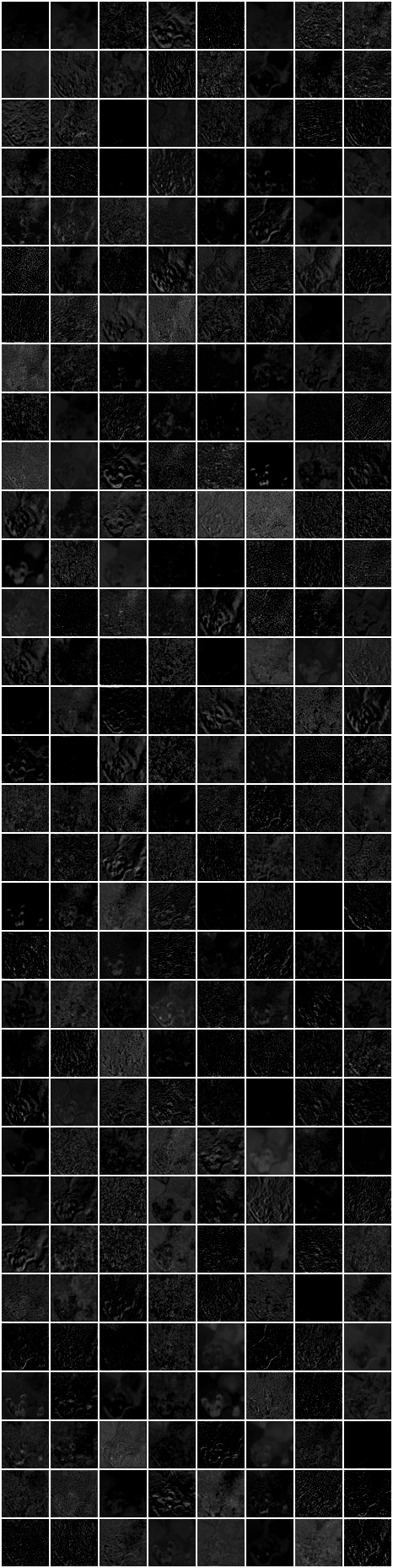}
\caption{Part of the feature maps output from Layer-1 of ResNet-50 with the ferret images as input}
\label{fig:resnet50-1}
\end{figure}

\begin{figure}[htb]
\centering
\includegraphics[trim=0 241 0 0,clip, width=0.99\linewidth]{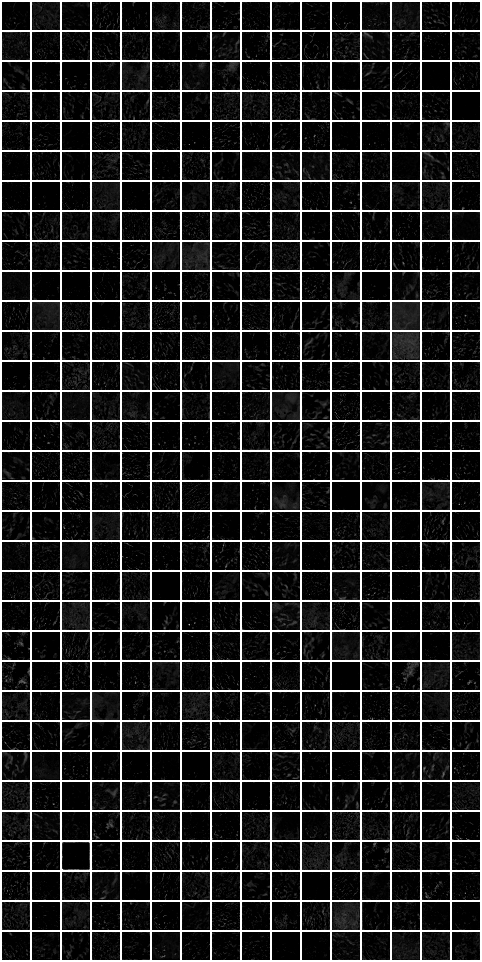}
\caption{All the feature maps output from Layer-2 of ResNet-50 with the ferret images as input}
\label{fig:resnet50-2}
\end{figure}

\begin{figure}[htb]
\centering
\includegraphics[trim=0 0 0 0,clip, width=0.99\linewidth]{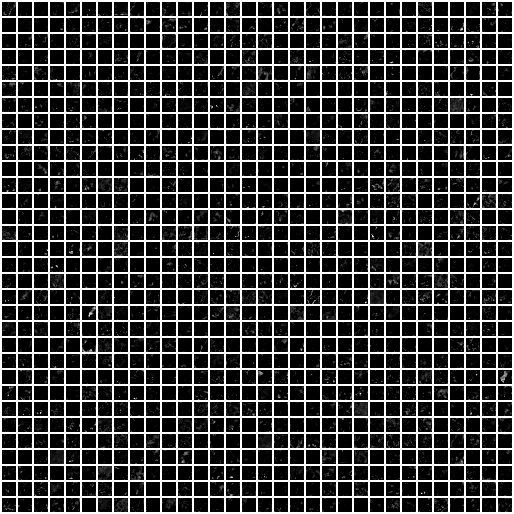}
\caption{All the feature maps output from Layer-3 of ResNet-50 with the ferret images as input}
\label{fig:resnet50-3}
\end{figure}

\begin{figure}[htb]
\centering
\includegraphics[trim=0 0 0 0,clip, width=0.99\linewidth]{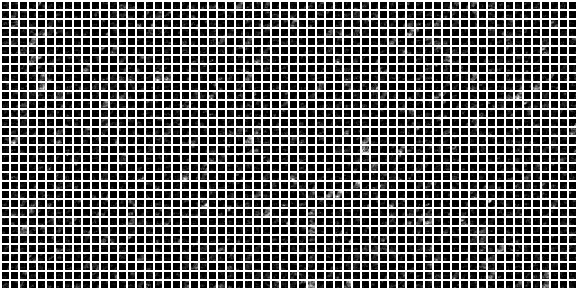}
\caption{All the feature maps output from Layer-4 of ResNet-50 with the ferret images as input}
\label{fig:resnet50-4}
\end{figure}

\begin{figure}[htb]
\centering
\includegraphics[trim=0 0 0 0,clip, width=0.99\linewidth]{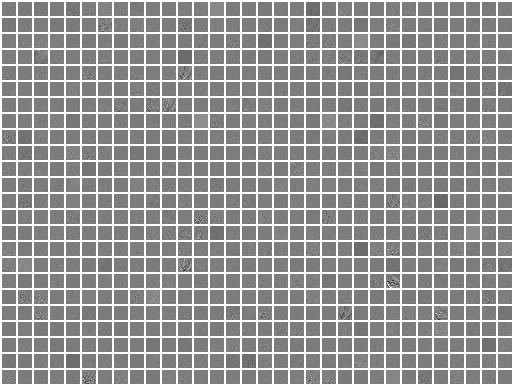}
\caption{All the feature maps output from Block-2 of Mixer-B/16 with the ferret images as input}
\label{fig:mixer-b16-1}
\end{figure}

\begin{figure}[htb]
\centering
\includegraphics[trim=0 0 0 0,clip, width=0.99\linewidth]{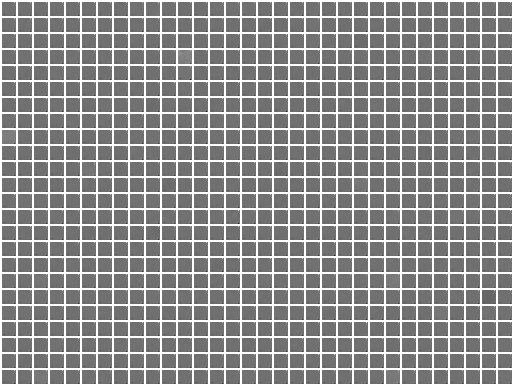}
\caption{All the feature maps output from Block-4 of Mixer-B/16 with the ferret images as input}
\label{fig:mixer-b16-2}
\end{figure}

\begin{figure}[htb]
\centering
\includegraphics[trim=0 0 0 0,clip, width=0.99\linewidth]{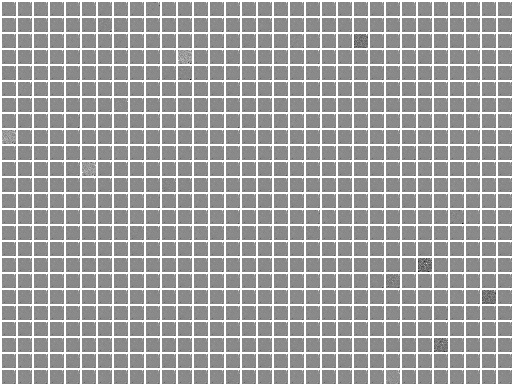}
\caption{All the feature maps output from Block-10 of Mixer-B/16 with the ferret images as input}
\label{fig:mixer-b16-3}
\end{figure}

\begin{figure}[htb]
\centering
\includegraphics[trim=0 0 0 0,clip, width=0.99\linewidth]{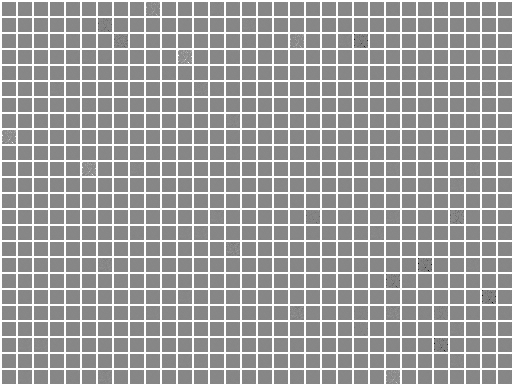}
\caption{All the feature maps output from Block-12 of Mixer-B/16 with the ferret images as input}
\label{fig:mixer-b16-4}
\end{figure}

\begin{figure}[htb]
\centering
\includegraphics[trim=0 0 0 0,clip, width=0.99\linewidth]{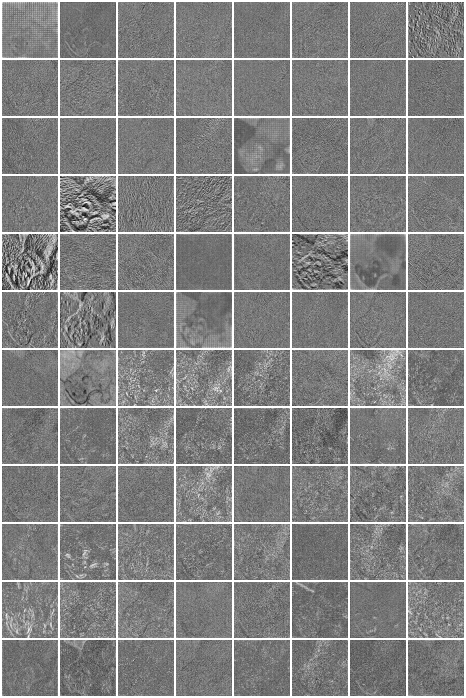}
\caption{All the feature maps output from Level-1 of RaftMLP-M with the ferret images as input}
\label{fig:raft-m-1}
\end{figure}

\begin{figure}[htb]
\centering
\includegraphics[trim=0 0 0 0,clip, width=0.99\linewidth]{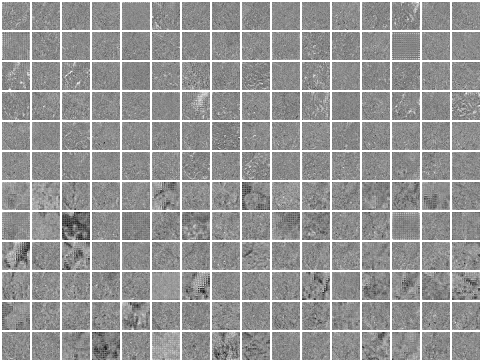}
\caption{All the feature maps output from Level-2 of RaftMLP-M with the ferret images as input}
\label{fig:raft-m-2}
\end{figure}

\begin{figure}[htb]
\centering
\includegraphics[trim=0 0 0 0,clip, width=0.99\linewidth]{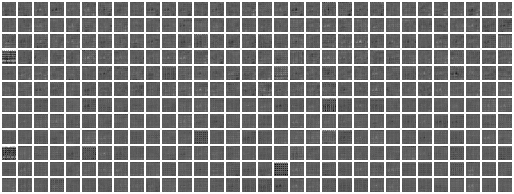}
\caption{All the feature maps output from Level-3 of RaftMLP-M with the ferret images as input}
\label{fig:raft-m-3}
\end{figure}

\begin{figure}[htb]
\centering
\includegraphics[trim=0 0 0 0,clip, width=0.99\linewidth]{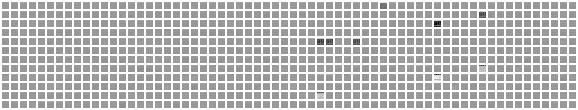}
\caption{All the feature maps output from Level-4 of RaftMLP-M with the ferret images as input}
\label{fig:raft-m-4}
\end{figure}

We used an image with ImageNet to visualize and compare its feature map. The image used as input was the ferret image on the left of Fig. \ref{fig:featuremaps}, which was input to pre-trained ResNet-50, Mixer-B/16, and RaftMLP-M. Some of the outputs of the intermediate layers are summarized on the right side of Fig. \ref{fig:featuremaps}. For ResNet-50, we used the output of layers 1 through 4; for Mixer-B/16, we used the output of Blocks 2, 4, 10, and 12; for RaftMLP-M, we used the output of each Level. We have also included further intermediate layer outputs for the three models, see Fig. \ref{fig:resnet50-1}, \ref{fig:resnet50-2}, \ref{fig:resnet50-3}, and \ref{fig:resnet50-4}
 for Resnet-50, Fig. \ref{fig:mixer-b16-1}, \ref{fig:mixer-b16-2}, \ref{fig:mixer-b16-3} and \ref{fig:mixer-b16-4} for Mixer-B/16, and Fig. \ref{fig:raft-m-1}, \ref{fig:raft-m-2}, \ref{fig:raft-m-3}, and \ref{fig:raft-m-4} for RaftMLP-M.
 
As mentioned in Section 5, the appearance of features in the middle layer of global MLP-based models is different from that of the convolutional base represented by ResNet. We believe this is why global MLP-based models do not perform well when selected as the backbone of existing architectures for object detection, instance segmentation, and semantic segmentation. The feature map of RaftMLP-M is different from that of ResNet in that the lower layers have feature maps that capture the features of the ferret. In contrast, the upper layers have feature maps with visible artifacts of vertical and horizontal lines. The feature maps of Mixer-B/16 do not capture the features of the ferret, and they are overall shuffled and have many similar feature maps. Tasks such as object detection, semantic segmentation, or even image generation will require innovations specific to global MLP-based models. The occurrence of artifacts might have a minor impact on classification, where global average pooling is used. However, for tasks such as segmentation and image generation, it becomes a severe problem. Hence, it will be necessary to design architectures and loss functions that do not emit this artifact. Or else, convolution-based methods such as RetinaNet, Mask R-CNN, and Semantic FPN may be insufficient to recover the whole shuffled information by global MLP-based models. To recover the global shuffled information by the global MLP-based model, global MLP-based models may lack a module that can capture the global relations, such as self-attention modules and token-mixing blocks.

\begin{table*}[htb]
\centering
\caption{Specific settings on the model architectures of hierarchy RaftMLP in different scales. $l$ denotes level and $c'_l$ denotes the number of basic channels in RaftMLP for level $l$.}
\begin{tabular}{c|cc|cc|cc@{}}
\toprule
\multicolumn{1}{c}{Model}&\multicolumn{2}{c}{RaftMLP-S}&\multicolumn{2}{c}{RaftMLP-M}&\multicolumn{2}{c}{RaftMLP-L} \\ \midrule
Level&Block&Setting&Block&Setting&Block&Setting \\ \midrule
$l=1$&RaftMLP$\times 2$&$c'_1=64$&RaftMLP$\times 2$&$c'_1=96$&RaftMLP$\times 2$&$c'_1=128$ \\
$l=2$&RaftMLP$\times 2$&$c'_2=128$&RaftMLP$\times 2$&$c'_2=192$&RaftMLP$\times 2$&$c'_1=192$\\
$l=3$&RaftMLP$\times 6$&$c'_3=256$&RaftMLP$\times 6$&$c'_3=384$&RaftMLP$\times 6$&$c'_1=512$\\
$l=4$&RaftMLP$\times 2$&$c'_4=512$&RaftMLP$\times 2$&$c'_4=768$&RaftMLP$\times 2$&$c'_1=1024$\\ \bottomrule
\end{tabular}
\label{table:models}
\end{table*}

\begin{table*}[ht]
\centering
\caption{Comparison of RetinaNet metrics trained on MS COCO with each ResNet, PureMLP, ConvMLP, RaftMLP, and Mixer as the backbone.}
\begin{tabular}{l|c|ccc|ccc@{}}
\toprule
Backbone & \#MParams & $AP^{b}$ & $AP^{b}_{50}$ & $AP^{b}_{75}$ & $AP^{b}_{S}$ & $AP^{b}_{M}$ & $AP^{b}_{L}$ \\
\midrule
ResNet-18~\cite{he2016deep,li2021convmlp} & 21.3 & 31.8 & 49.6 & 33.6 & 16.3 & 34.3 & 43.2 \\
PureMLP-S~\cite{li2021convmlp} & 17.6 & 27.1 & 44.2 & 28.3 & 13.6 & 29.2 & 36.4 \\
ConvMLP-S~\cite{li2021convmlp} & 18.7 & 37.2 & 56.4 & 39.8 & 20.1 & 40.7 & 50.4 \\
RaftMLP-S & 19.6 & 17.7 & 33.3 & 16.5 & 4.5 & 14.1 & 32.4 \\
\midrule
ResNet-50~\cite{he2016deep,li2021convmlp} & 37.7 & 36.3 & 55.3 & 38.6 & 19.3 & 40.0 & 48.8 \\
PureMLP-M~\cite{li2021convmlp} & 25.9 & 28.0 & 45.6 & 29.0 & 14.5 & 29.9 & 37.8 \\
ConvMLP-M~\cite{li2021convmlp} & 27.1 & 39.4 & 58.7 & 42.0 & 21.5 & 43.2 & 52.5 \\
RaftMLP-M & 27.1 & 19.3 & 36.3 & 17.8 & 5.2 & 15.9 & 35.1 \\
\midrule
ResNet-101~\cite{he2016deep,li2021convmlp} & 56.7 & 38.5 & 57.8 & 41.2 & 21.4 & 42.6 & 51.1 \\
PureMLP-L~\cite{li2021convmlp} & 50.1 & 28.8 & 46.8 & 29.9 & 15.0 & 31.0 & 38.4 \\
ConvMLP-L~\cite{li2021convmlp} & 52.9 & 40.2 & 59.3 & 43.3 & 23.5 & 43.8 & 53.3 \\
RaftMLP-L & 52.9 & 19.5 & 36.8 & 18.1 & 5.0 & 16.1 & 35.4 \\
\midrule
Mixer-B/16~\cite{tolstikhin2021mlp} & 70.3 & 10.7 & 20.0 & 10.1 & 0.1 & 6.7 & 25.8 \\
\bottomrule
\end{tabular}
\label{table:retinanet}
\end{table*}

\begin{table*}[!ht]
\centering
\caption{Comparison of Mask R-CNN metrics trained on MS COCO with each ResNet, PureMLP, ConvMLP, RaftMLP and Mixer as the backbone.}
\begin{tabular}{l|c|ccc|ccc@{}}
\toprule
Backbone & \#MParams & $AP^{b}$ & $AP^{b}_{50}$ & $AP^{b}_{75}$ & $AP^{m}$ & $AP^{m}_{50}$ & $AP^{m}_{75}$ \\
\midrule
ResNet-18~\cite{he2016deep,li2021convmlp} & 31.2 & 34.0 & 54.0 & 36.7 & 31.2 & 51.0 & 32.7 \\
PureMLP-S~\cite{li2021convmlp} & 27.5 & 25.1 & 45.1 & 25.1 & 25.0  & 42.8  & 26.0 \\
ConvMLP-S~\cite{li2021convmlp} & 28.7 & 38.4 & 59.8 & 41.8 & 35.7 & 56.7 & 38.2 \\
RaftMLP-S & 29.5 & 21.8 & 40.2 & 21.0 & 19.7 & 36.5 & 19.1 \\
\midrule
ResNet-50~\cite{he2016deep,li2021convmlp} & 44.2 & 38.0 & 58.6 & 41.4 & 34.4 & 55.1 & 36.7 \\
PureMLP-M~\cite{li2021convmlp} & 35.8 & 25.8 & 46.1 &25.8 & 25.6 & 43.5 & 26.5 \\
ConvMLP-M~\cite{li2021convmlp} & 37.1 & 40.6 & 61.7 & 44.5 & 37.2 & 58.8 & 39.8 \\
RaftMLP-M & 40.9 & 23.4 & 42.5 & 22.7 & 21.1 & 38.8 & 20.8 \\
\midrule
ResNet-101~\cite{he2016deep,li2021convmlp} & 63.2 & 40.4 & 61.1 & 44.2 & 36.4 & 57.7 & 38.8 \\
PureMLP-L~\cite{li2021convmlp} & 59.5  & 26.5 & 45.0 &27.4 & 26.7 & 47.5 & 26.8 \\
ConvMLP-L~\cite{li2021convmlp} & 62.2 & 41.7 & 62.8 & 45.5 & 38.2 & 59.9 & 41.1 \\
RaftMLP-L & 55.5 & 24.2 & 43.9 & 23.7 & 21.6 & 39.7 & 23.7 \\
\midrule
Mixer-B/16~\cite{tolstikhin2021mlp} & 79.8 & 11.9 & 22.8 & 11.2 & 9.5 & 19.1 & 8.5 \\
\bottomrule
\end{tabular}
\label{table:maskrcnn}
\end{table*}

\begin{table}[!ht]
\centering
\caption{Comparison of Semantic FPN metrics trained on MS COCO with each ResNet, PureMLP, ConvMLP, RaftMLP and Mixer as the backbone.}
\begin{tabular}{l|c|c@{}}
\toprule
Backbone & \#MParams & mIoU \\
\midrule
ResNet-18~\cite{he2016deep,li2021convmlp} & 15.5 & 32.9 \\
Pure-MlP-S~\cite{li2021convmlp} & 11.6 & 23.9 \\
ConvMLP-S~\cite{li2021convmlp} & 12.8 & 35.8 \\
RaftMLP-S & 13.6 & 30.7 \\
\midrule
ResNet-50~~\cite{he2016deep,li2021convmlp} & 28.5 & 36.7 \\
Pure-MlP-M~\cite{li2021convmlp} & 19.9 & 25.2 \\
ConvMLP-M~\cite{li2021convmlp} & 21.1 & 38.6  \\
RaftMLP-M & 25.0 & 32.3 \\
\midrule
ResNet-101~\cite{he2016deep,li2021convmlp} & 47.5 & 38.8 \\
Pure-MlP-L~\cite{li2021convmlp} & 43.6 & 26.3 \\
ConvMLP-L~\cite{li2021convmlp} & 46.3 & 40.0 \\
RaftMLP-L & 39.6 & 33.3 \\
\midrule
Mixer-B/16~\cite{tolstikhin2021mlp} & 63.9 & 28.1 \\
\bottomrule
\end{tabular}
\label{table:semseg}
\end{table}




%
%
\bibliographystyle{splncs04}
\bibliography{egbib}